\documentclass[runningheads]{llncs}

\usepackage[mobile]{eccv}

\usepackage{eccvabbrv}
\usepackage[framemethod=TikZ]{mdframed}

\usepackage{graphicx}
\usepackage{booktabs}
\usepackage{float}
\usepackage[table,xcdraw]{xcolor}
\usepackage[accsupp]{axessibility}  

\usepackage{hyperref}

\usepackage{orcidlink}
\usepackage{makecell}
\usepackage{cite} 

\usepackage{capt-of}
\usepackage{graphicx}
\usepackage{lipsum}
\usepackage{pifont}
\usepackage{pgffor}
\usepackage{nicematrix}
\usepackage{xcolor}

\usepackage{booktabs}
\usepackage{xcolor}
\usepackage{lipsum} 
\usepackage{blindtext}
\usepackage[labelsep=period]{caption}
\usepackage{floatflt}
\usepackage{caption}
\usepackage[framemethod=TikZ]{mdframed} 
\usepackage[most]{tcolorbox}
\usepackage{placeins}   
\usepackage{wrapfig}
\usepackage{tabularx}

\definecolor{headergray}{gray}{0.9}
\definecolor{audioblue}{rgb}{0.0, 0.4, 0.7}
\newcommand{\cmark}{\ding{51}} 
\newcommand{\xmark}{\ding{55}} 

\newcommand{\benchmarkname}{E3VS-Bench}

\begin{document}

 \title{E3VS-Bench: A Benchmark for Viewpoint-Dependent Active Perception in 3D Gaussian Splatting Scenes} 

\titlerunning{E3VS-Bench}

\author{Koya Sakamoto\inst{1} \and
Taiki Miyanishi\inst{1} \and
Daichi Azuma\inst{1} \and
Shuhei Kurita\inst{2,3,4} \and \\
Shu Morikuni\inst{1} \and
Naoya Chiba\inst{5} \and
Motoaki Kawanabe\inst{6} \and \\ 
Yusuke Iwasawa\inst{1} \and
Yutaka Matsuo\inst{1}}

\authorrunning{K.~Sakamoto et al.}

\institute{The University of Tokyo, Japan \and
National Institute of Informatics, Japan \and
Institute of Science Tokyo, Japan  \and
NII LLMC, Japan \and
The University of Osaka, Japan \and
Advanced Telecommunications Research Institute International, Japan}
\vspace{-2em}

\maketitle

\begin{abstract}
Visual search in 3D environments requires embodied agents to actively explore their surroundings and acquire task-relevant evidence. 
However, existing visual search and embodied AI benchmarks, including EQA, typically rely on static observations or constrained egocentric motion, and thus do not explicitly evaluate fine-grained viewpoint-dependent phenomena that arise under unrestricted 5-DoF viewpoint control, such as disambiguating object attributes observable only from specific angles.
To address this limitation, we introduce {E3VS-Bench}, a benchmark for embodied 3D visual search where agents must control their viewpoints in 5-DoF to gather viewpoint-dependent evidence for question answering. 
E3VS-Bench consists of 99 high-fidelity 3D scenes reconstructed using 3D Gaussian Splatting and 2,014 question-driven episodes.
3D Gaussian Splatting enables photorealistic free-viewpoint rendering that preserves fine-grained visual details (e.g., small text and subtle attributes) often degraded in mesh-based simulators, thereby allowing the construction of questions that cannot be answered from a single view and instead require active inspection across viewpoints in 5-DoF.
We evaluate multiple state-of-the-art VLMs and compare their performance with humans.
Despite strong 2D reasoning ability, all models exhibit a substantial gap from humans, highlighting limitations in active perception and coherent viewpoint planning specifically under full 5-DoF viewpoint changes.
The benchmark, code, and dataset are publicly available at \url{https://k0uya.github.io/e3vs-proj/}.

\end{abstract}    
\begin{figure}[t]
\centering
\includegraphics[width=\linewidth]{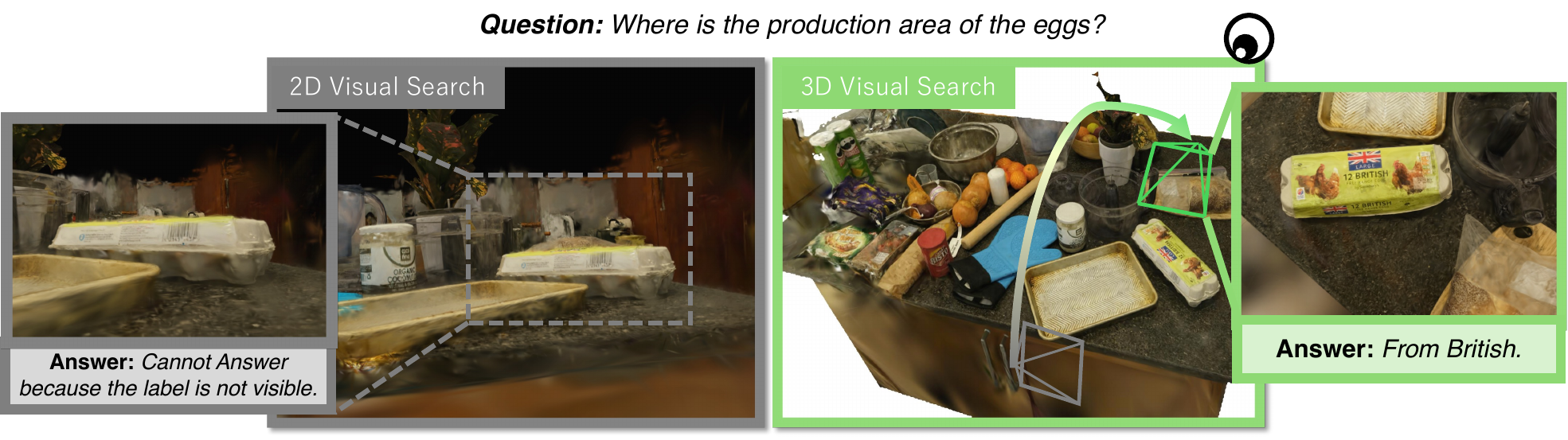}
\vspace{-2.0em}
\caption{
Overview of the proposed Embodied 3D Visual Search (E3VS) task. 
Unlike 2D visual search, E3VS requires an agent to actively control its 5-DoF viewpoint to resolve occlusions and acquire fine-grained visual evidence, such as the production area label on an egg carton.
}
\vspace{-1.5em}
\label{fig:teaser}
\end{figure}

\vspace{-1.5em}
\section{Introduction}

Active perception refers to an agent's ability to actively control its sensory apparatus to acquire task-relevant information, rather than passively receiving observations~\cite{Aloimonos1988ActiveVision}.
In modern embodied AI, it enables agents to selectively gather task-relevant information by controlling their viewpoints and exploring the environment~\cite{Feng2025EmbodiedAI}.
Such active control is essential for understanding 3D structure and for acquiring information that cannot be obtained from a single static observation.
However, existing benchmarks and agents remain largely constrained to two-dimensional behaviors~\cite{Qi2020reverie,eqamatterport}, and therefore do not fully capture or evaluate these capabilities in real-world 3D environments.

Visual search has long been studied in psychology and neuroscience as the process of locating targets through goal-directed and saliency-driven attention~\cite{wolfe2020visual,wolfe2011search,torralba2006contextual}.
Inspired by these studies, recent computer vision works have explored visual search using multimodal models~\cite{Wu2024vstar,dyfo2025,yu2025thinking360deghumanoidvisual}.
However, their evaluation remains confined to static 2D observations and cannot capture spatial reasoning challenges in real 3D environments, such as occlusions, geometric relations, and depth ambiguity.
Within embodied AI, Embodied Question Answering (EQA) requires an agent to navigate a 3D environment to gather visual evidence and answer a question~\cite{abhishek2018eqa,shridhar2020alfred}.
However, most existing EQA settings treat the agent as a 2D moving camera restricted to planar navigation and limited rotation, and thus do not evaluate the viewpoint control required for realistic 3D perception.
In real-world environments, a target may be visible from the initial viewpoint, yet its identifying attributes, such as labels or fine-grained text, remain unreadable without actively adjusting the viewpoint.
This type of viewpoint-dependent evidence remains largely unexplored in existing benchmarks.

While recent tasks such as Aerial VLN~\cite{Liu2023AerialVLN,lee2025citynav} extend navigation into 3D space, their primary focus is large-scale trajectory following toward distant goals.
They do not evaluate the fine-grained viewpoint control required for active inspection.
In these situations, the ability to manipulate the 5-DoF viewpoint itself becomes the key to revealing hidden information.

To address this gap, we introduce \textbf{E3VS-Bench}, a benchmark for Embodied 3D Visual Search (E3VS), where agents must actively control their viewpoints in 3D space with full 5-DoF to acquire task-relevant visual evidence for question answering.
The benchmark contains 99 high-fidelity reconstructed scenes and 2,014 question-driven episodes.
E3VS-Bench is built upon publicly available scenes reconstructed using 3D Gaussian Splatting (3DGS)~\cite{kerbl2023GaussianSplatting}.
3DGS enables photorealistic free-viewpoint rendering that preserves subtle details, such as fine-grained text and brand logos, often degraded in mesh-based representations. 
This fidelity makes it possible to define viewpoint-dependent questions that cannot be answered from a single observation and require active viewpoint exploration.

To ensure that benchmark performance reflects genuine exploration ability rather than prior knowledge or favorable initial viewpoints, we introduce a rigorous episode filtering pipeline based on a VLM-as-a-judge framework.
This process removes episodes that can be solved without meaningful viewpoint transitions.

Our main contributions are summarized as follows:
\begin{itemize}
    \item We study \textbf{E3VS}, a setting where agents must acquire task-relevant visual evidence through 5-DoF viewpoint control.
    \item We present \textbf{E3VS-Bench}, a benchmark built on 99 photorealistic 3D Gaussian Splatting scenes, with 2,014 human-annotated episodes for evaluating viewpoint-dependent reasoning and active perception in 3D environments.
    \item We conduct a comprehensive evaluation of state-of-the-art VLMs and reveal that current models 
     struggle with active viewpoint planning and exploration despite strong performance on conventional 2D reasoning tasks.    
\end{itemize}
\section{Related Work}
\label{sec:relatedwork}
\noindent \textbf{Visual Search.}
Classical visual search studies in psychology and neuroscience~\cite{wolfe2020visual, wolfe2011search, torralba2006contextual} investigate how humans locate target objects among distractors through goal-directed and saliency-driven attention.
Inspired by these findings, recent computer vision studies have explored visual search using deep and multimodal models~\cite{pixelreasoner2025, Wu2024vstar, dyfo2025, thyme2025}, employing mechanisms such as curiosity-driven reinforcement learning in pixel space (Pixel Reasoner~\cite{pixelreasoner2025}), dynamic focus (SEAL~\cite{Wu2024vstar}, DyFo~\cite{dyfo2025}), and reinforcement-based fine-tuning (Thyme~\cite{thyme2025}).
To evaluate such capabilities, several benchmarks have been proposed, including Mini-O3~\cite{lai2025mini-o3}, H*Bench~\cite{yu2025thinking360deghumanoidvisual}, and O3-Bench~\cite{li2026insight-o3}.
Mini-O3 collects challenging visual search problems requiring exploratory reasoning, while H*Bench simulates pseudo-actions such as \texttt{turn\_left} and \texttt{turn\_right} using panoramic images.
O3-Bench requires models to reference multiple image regions during reasoning.
These environments, however, remain limited to static 2D imagery and cannot evaluate occlusion reasoning or spatial relationships in 3D environments.

\vspace{0.1cm}
\noindent \textbf{Active Perception and Exploration.}
Building upon the concept of active perception~\cite{Bajcsy1988ActivePerception, Aloimonos1988ActiveVision}, recent embodied AI studies investigate how agents adjust viewpoints to obtain informative observations, through active perception for object understanding and manipulation~\cite{embodied_amodal, xiong2025via, kerrj2025eyerobot} or exploration-driven strategies that seek novel or semantically meaningful observations~\cite{chaplot2020semantic, lookaround2024, tenny2025womap}. 
Most of these methods focus on specific objectives such as object detection or navigation.
Building upon this line of work, our setting studies how agents actively acquire task-relevant visual evidence through fine-grained 5-DoF viewpoint control for language-guided reasoning.

\vspace{0.1cm}
\noindent \textbf{Question Answering in 3D Scene.}
Question answering in 3D environments follows two main paradigms: scene-centric 3D-QA and embodied QA.
Scene-centric approaches reason directly over reconstructed 3D representations without requiring physical movement.
For example, ScanQA~\cite{Azumascanqa} extends the ScanRefer~\cite{chen2020scanrefer} paradigm from 3D object grounding to QA on point-cloud scenes, and SQA3D~\cite{ma2022sqa3d} introduces situated reasoning from a specific pose using pre-defined observations.
These approaches assume access to complete scene representations and thus do not evaluate how agents actively acquire missing visual evidence.
EQA instead requires an agent to navigate a 3D environment to answer natural-language questions~\cite{abhishek2018eqa, eqamatterport, mahumar2024openeqa, sakamoto2024mapeqa}. 
Answerability Fields~\cite{azuma2024answerability} estimates spatial answerability distributions, predicting locations where visual evidence required to answer a question is likely to be observable. Subsequent studies enhance relational reasoning through scene graphs and long-horizon planning~\cite{grapheqa2025, efficienteqa2025, enter2025}.
Simulation platforms such as Habitat~\cite{savva2019habitat} enable benchmarks including MP3D-EQA~\cite{eqamatterport} and EXPRESS-BENCH~\cite{Jiang2025express-bench-eqa}.
These navigation-centric settings, however, often treat the agent as a 2D moving camera and do not evaluate the fine-grained viewpoint control required to reveal task-relevant visual details.

\begin{table}[t]
\centering
\renewcommand{\arraystretch}{0.95}
\resizebox{\linewidth}{!}{
\begin{tabular}{lccccc}
\toprule
\textbf{Dataset} 
& \textbf{Active} & \textbf{Env.} & \textbf{Source} & \textbf{DoF} & \textbf{Episodes} \\
\midrule
ScanQA~\cite{Azumascanqa} & \xmark & Mesh & ScanNet~\cite{scannetdata} & -- & 41,363 \\
SQA3D~\cite{ma2022sqa3d} & \xmark & Mesh & ScanNet~\cite{scannetdata}  & -- & 33,400 \\
REVERIE~\cite{Qi2020reverie} & \cmark & Mesh & Matterport3D~\cite{matterport2017} & 2-DoF (Graph-based) & 21,702 \\
EQA~\cite{abhishek2018eqa} & \cmark & Mesh & House3D~\cite{wu2018house3d} & 2-DoF & 5,281 \\
MP3D-EQA~\cite{eqamatterport} & \cmark & Mesh & Matterport3D~\cite{matterport2017} & 2-DoF & 1,136 \\
OpenEQA~\cite{mahumar2024openeqa} & \cmark & Mesh & Matterport3D~\cite{matterport2017} & 2-DoF & 1,600 \\
\midrule
\textbf{E3VS-Bench} & \textbf{\cmark} & \textbf{3D-GS} & \textbf{SceneSplat++}~\cite{ma2025scenesplat++} & \textbf{5-DoF} & 2,014 \\
\bottomrule
\end{tabular}
}
\caption{
Comparison with embodied QA and 3D visual grounding datasets.
E3VS-Bench uses 3DGS scenes with 5-DoF control for occlusion-aware active perception.
}
\label{tab:related_datasets}
\vspace{-2.25em}
\end{table}

Table~\ref{tab:related_datasets} summarizes major 3D scene QA and EQA datasets.
Although these datasets vary in scale and realism, they typically rely on mesh or point-cloud reconstructions that restrict viewpoint interaction and limit photometric fidelity, and thus do not evaluate how viewpoint-dependent observations influence question answering.
In contrast, E3VS-Bench enables agents to actively explore photorealistic 3DGS environments with full 5-DoF viewpoint control, allowing systematic evaluation of viewpoint-dependent reasoning in 3D environments.

\vspace{-0.5cm}
\section{Benchmark Design and Construction}~\label{3_dataset_benchmark}
\vspace{-0.5cm}

Unlike existing embodied tasks (e.g., EQA~\cite{eqamatterport}) that rely on limited camera motion, E3VS-Bench requires agents to select viewpoints revealing task-relevant evidence not observable from a single view. By embedding such viewpoint-dependent constraints in photorealistic 3D scenes with unrestricted camera control, it evaluates viewpoint planning and occlusion-aware spatial reasoning.

\subsection{Task Formulation}
\label{sec:task_formulation}
\vspace{-0.1cm}
\noindent \textbf{Problem Setup.}
We formulate E3VS as an active perception task for question answering in photorealistic 3D environments.
Unlike conventional 2D visual search~\cite{Wu2024vstar, dyfo2025} or EQA~\cite{eqamatterport}, where agents typically move on a ground plane with limited viewpoint control, E3VS requires agents to adjust their viewpoints in 3D space to resolve occlusions, depth ambiguities, and fine-grained visual details.

Formally, an episode is defined by a triplet $(\mathcal{S}, q, v_0)$, where $\mathcal{S}$ denotes a 3D scene reconstructed using 3DGS, $q$ is a natural-language question, and $v_0$ represents the initial camera viewpoint.
At each discrete time step $t \in [0, T]$, the agent receives an egocentric RGB observation $O_t \in \mathbb{R}^{H \times W \times 3}$ rendered from the scene at viewpoint $v_t$.
The initial viewpoint $v_0$ is centered on the target object; however, the target may be partially or fully occluded by other objects.

\vspace{0.1cm}
\noindent \textbf{Action Space.}
We represent the agent's state at time $t$ as a viewpoint 
$v_t = (x_t, y_t, z_t, \theta_t, \phi_t) \in \mathbb{R}^5$,
where $(x_t, y_t, z_t)$ denote the 3D Cartesian coordinates and 
$(\theta_t, \phi_t)$ represent the yaw and pitch angles.
The agent interacts with the environment through a discrete 5-DoF action space $\mathcal{A}$, where each action $a_t \in \mathcal{A}$ corresponds to either a translation, a rotation, or termination.
The state transition follows the environment dynamics $v_{t+1} = E(v_t, a_t)$, where a collision after executing $a_t$ leaves the state unchanged $v_{t+1} = v_t$.

\vspace{0.1cm}
\noindent \textbf{Success Criteria.}
Each episode is associated with a set of human-annotated answerable viewpoints $\mathcal{V}_{ans}$ and a corresponding goal image $O_{goal}$ rendered from $v \in \mathcal{V}_{ans}$.
Upon executing the \texttt{stop} action at step $T$, the agent provides a predicted answer $\hat{y}$ based on the final observation $O_T$ and the question $q$. 

To evaluate the correctness of open-vocabulary responses, we employ a {VLM-as-a-judge} framework, following LLM-Match in OpenEQA~\cite{mahumar2024openeqa}.
Specifically, an evaluator model $J$ (e.g., \textit{GPT 5.1}) receives the agent's final observation $O_T$, the question $q$, the predicted answer $\hat{y}$, the ground-truth answer $y$, and the goal image $O_{goal}$ to compute a score $S = J(O_T, q, \hat{y}, y, O_{goal}) \in \{1, \dots, 5\}$,
where 5 is perfectly correct and 1 entirely incorrect.
This metric allows for a flexible yet rigorous assessment of the agent's active perception capability.

\begin{figure}[t]
\centering
\includegraphics[width=0.9\columnwidth]{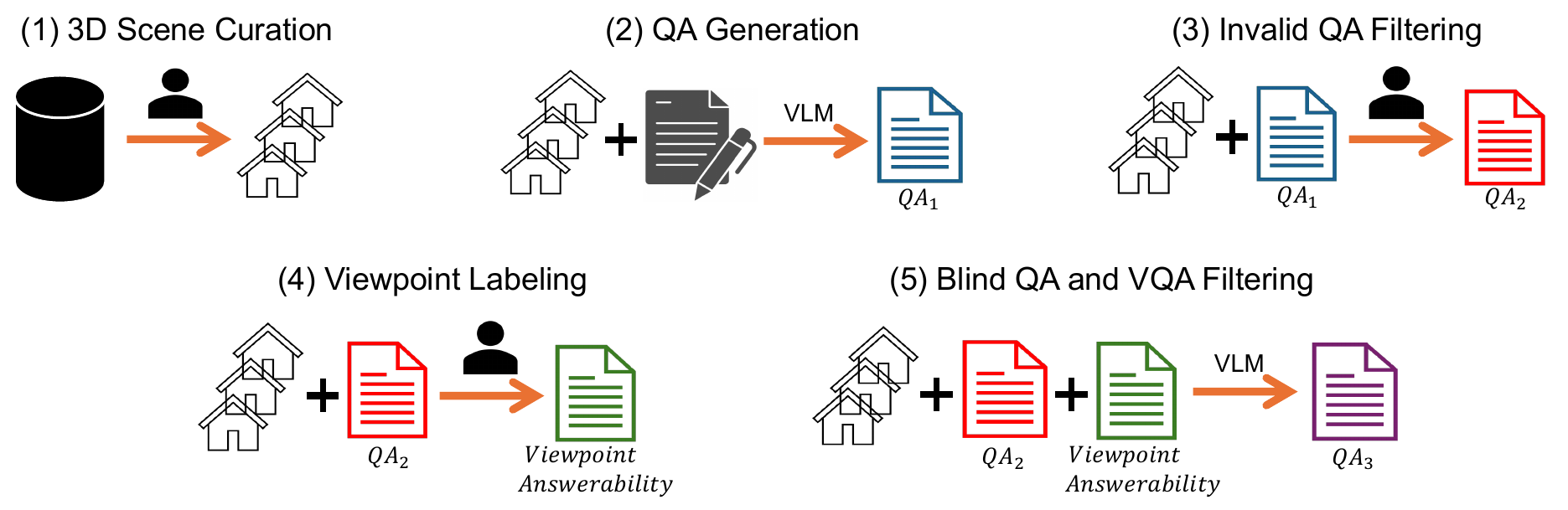}
\vspace{-1.0em}
\caption{
Dataset construction pipeline for E3VS-Bench.
The pipeline consists of five stages:
(1) 3D scene curation from SceneSplat++~\cite{ma2025scenesplat++},
(2) QA generation using a VLM,
(3) invalid QA filtering with human verification,
(4) viewpoint labeling to identify answerable viewpoints, and
(5) answerability filtering to remove questions solvable without viewpoint transitions.
}
\vspace{-1.5em}
\label{fig:dataset_construction}
\end{figure}

\subsection{Dataset Construction Pipeline}
\label{sec:data_construction_pipeline}
In this section, we present an overview of the question--answer creation pipeline used to construct \benchmarkname.
As shown in Fig.~\ref{fig:dataset_construction}, the construction process consists of five stages:
(1) \textit{3D scene curation},
(2) \textit{QA generation},
(3) \textit{invalid QA filtering},
(4) \textit{viewpoint labeling}, and
(5) \textit{answerability filtering}
, followed by additional data cleaning. We briefly describe each stage below and provide full details in the supplementary.

\begin{wrapfigure}{r}{0.4\linewidth}
\vspace{-20pt}
\centering
\includegraphics[width=\linewidth]{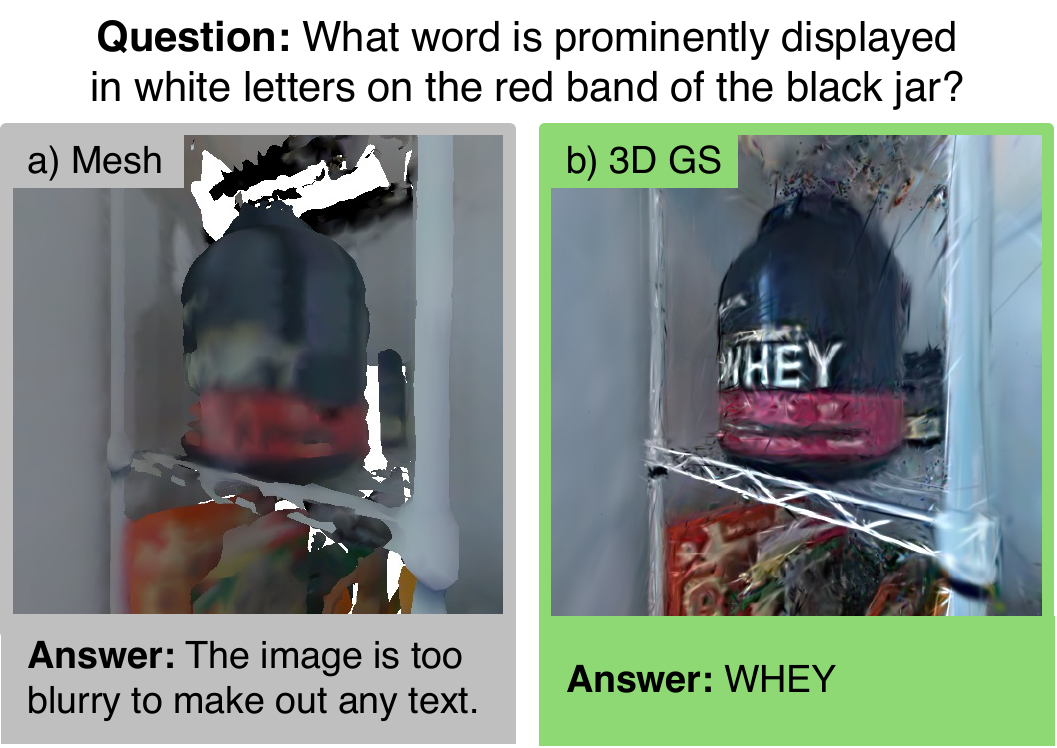}
\caption{
Comparison of rendering quality between traditional mesh-based (ScanNet++) and 3D Gaussian Splatting (SceneSplat++). 3DGS preserves sharp textures for small text (e.g., "WHEY" label), which is crucial for viewpoint-dependent visual reasoning.
}
\label{fig:comparison_mesh_3dgs}
\vspace{-10pt}
\end{wrapfigure}

\vspace{0.1cm}
\noindent \textbf{3D Scene Curation.}
First, we manually curate 105 high quality reconstructed scenes from 
{SceneSplat++}~\cite{ma2025scenesplat++},
which represents 3D environments from {ScanNet++}~\cite{yeshwanthliu2023scannetpp} using 3DGS~\cite{kerbl2023GaussianSplatting}.
Unlike traditional mesh-based representations that often suffer from texture degradation and geometric oversmoothing, 3DGS preserves the photometric fidelity necessary for reliable viewpoint-dependent reasoning.
As shown in Fig.~\ref{fig:comparison_mesh_3dgs}, traditional meshes often lack the photorealistic textures needed to capture subtle visual attributes, such as text on a plastic bag or brand logos. 
The photorealistic rendering of 3DGS enables fine-grained, viewpoint-dependent evaluation of embodied agent perception.

\vspace{0.1cm}
\noindent \textbf{QA Generation.}
To generate high-quality question–answer pairs at scale, we employ a three-stage pipeline powered by a VLM, specifically \textit{Gemini 2.5 Flash}.
(1) We select object categories with few instances per scene.
For each instance, we compute a viewing distance so that its projected object bounding box spans roughly three-fifths of the image along either axis.
We then uniformly sample viewpoints on a sphere centered at the object and render the corresponding multi-view images.
(2) We use a VLM to filter the generated views, discarding images where the object is heavily occluded, outside the reconstruction bounds, or visually ambiguous.
(3) Using the remaining valid multi-view images, the VLM synthesizes candidate question-answer pairs that cover a spectrum of visual reasoning, including intrinsic attributes (e.g., color, material, and branding) and extrinsic spatial relations.
This pipeline generates 27,877 QAs across 7,578 instances.

\vspace{0.1cm}
\noindent \textbf{Invalid QA Filtering.}
We perform human filtering of QA candidates and annotate viewpoint answerability. Annotators verify each candidate against predefined criteria (see Appendix), removing ambiguous, multi-interpretable, or objectively unanswerable questions. 
Inaccurate answers are corrected accordingly.
The overall human annotation process required approximately 1,120 hours.

\vspace{0.1cm}
\noindent \textbf{Viewpoint Labeling.}
For each retained question, expert annotators select viewpoints that contain sufficient information to answer it.
Physically invalid viewpoints, such as those intersecting 3D geometry, are discarded.
The annotations are further verified by another annotator to ensure quality and consistency.
Finally, unanswerable viewpoints serve as episode starting positions, while answerable viewpoints serve as goal positions.

\vspace{0.1cm}
\noindent \textbf{Answerability Filtering.}
To ensure the benchmark properly evaluates active exploration, we remove questions answerable from prior knowledge alone and episodes solvable from the initial viewpoint.
For filtering, we use \textit{GPT-5.1}. We perform VQA in both a blind setting (without visual input) and from the designated start viewpoint, evaluating response correctness on a 1--5 scale via a VLM-as-a-Judge protocol (see Section~\ref{sec:task_formulation} for details).
QA pairs scoring 3 or above in either setting are excluded, where the threshold is chosen as the midpoint of the 1--5 scale to conservatively filter out cases that show any indication of being answerable without active exploration.
The remaining episodes cannot be solved from prior knowledge or the initial observation alone, and require active 5-DoF viewpoint transitions and information gathering for accurate answers.

\subsection{Dataset Statistics and Analysis}
\begin{figure}[t]
\centering
\includegraphics[width=\columnwidth]{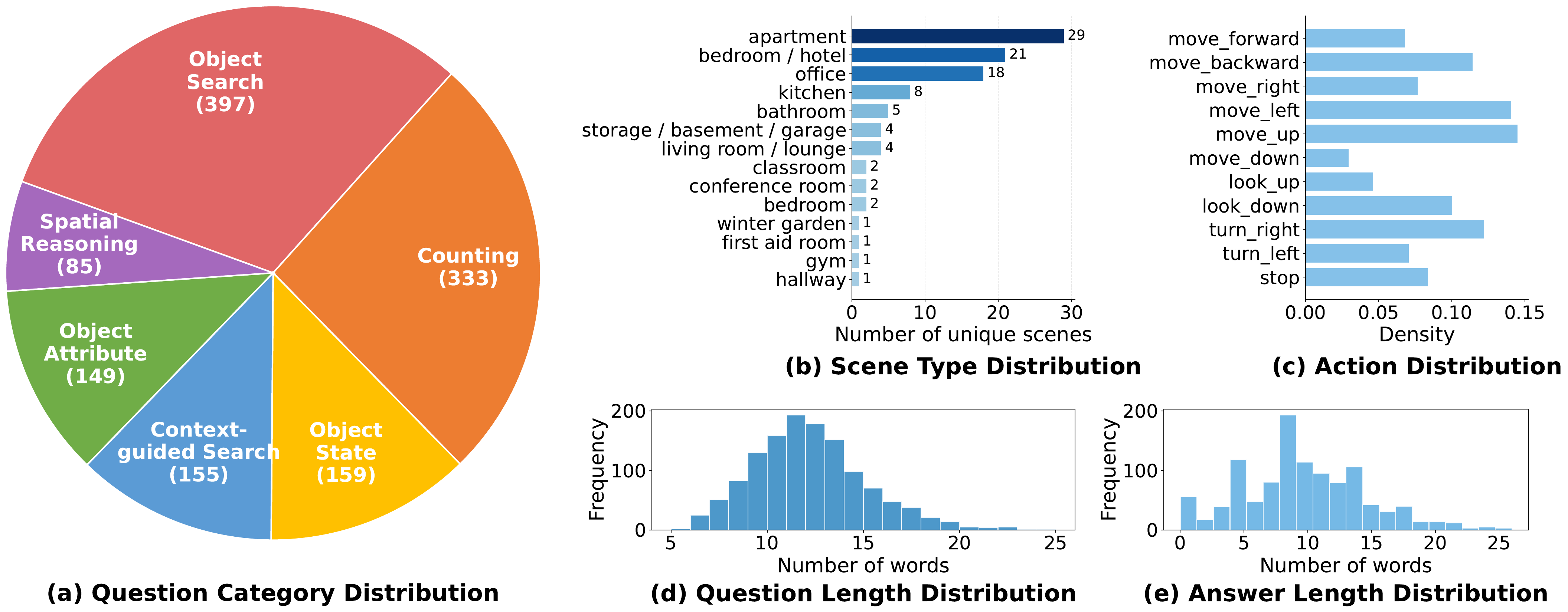}
\vspace{-1.5em}
\caption{
\textbf{Dataset Distribution.}
Each figure represents (a) question category distribution of unique QA pairs classified by \textit{GPT 5.1}. (b) scene type distribution. (c) action distribution. (d, e) the number of words in each question and answer.
}
\vspace{-1.25em}
\label{fig:dataset_statistics}
\end{figure}

\begin{figure}[t]
\centering
\includegraphics[width=\textwidth]{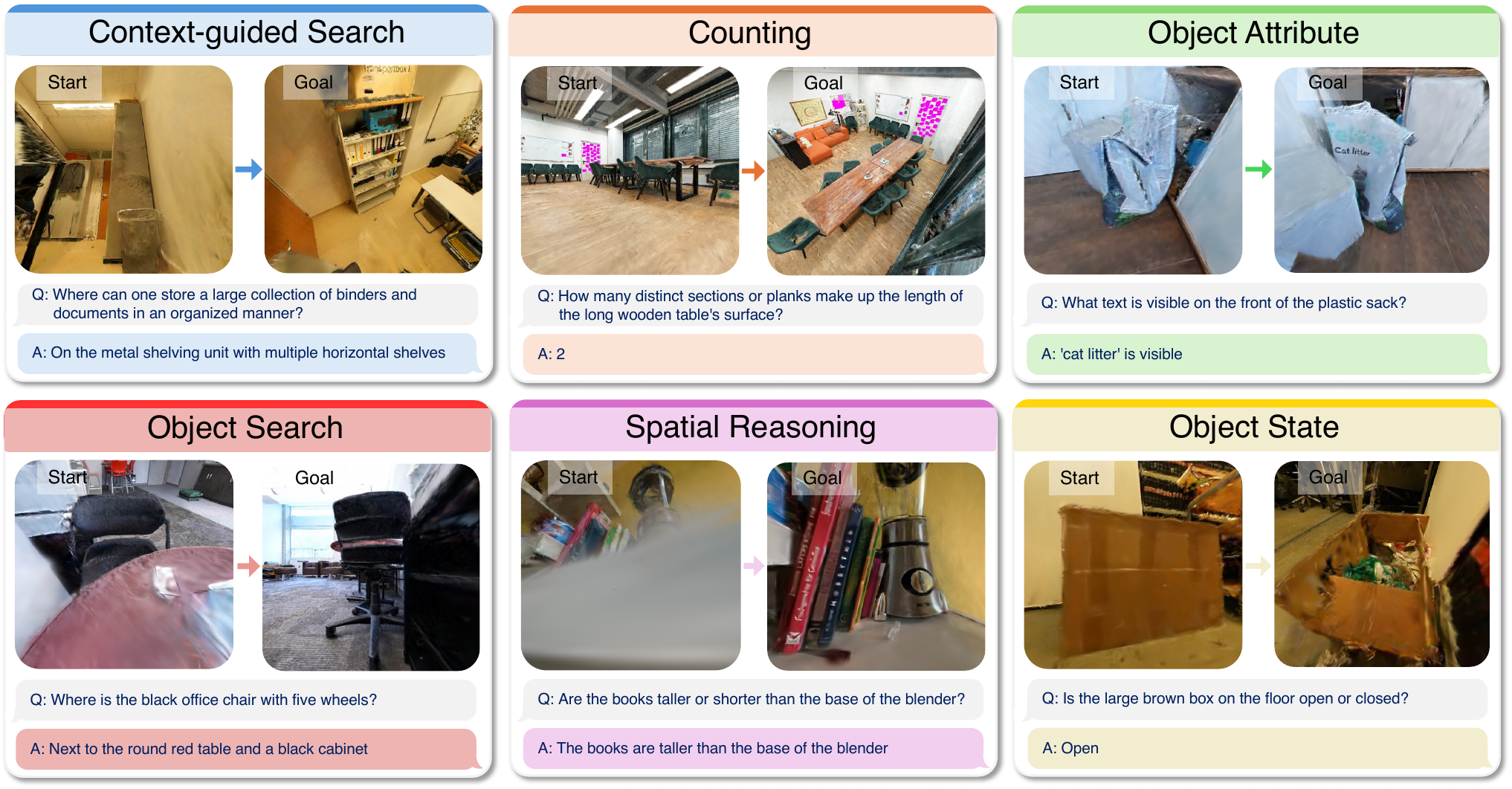}
\vspace{-1.5em}
\caption{Examples of E3VS task defined in our dataset.
Each example illustrates a distinct reasoning type that requires viewpoint control in reconstructed 3D environments.}
\vspace{-1.5em}
\label{fig:dataset_samples}
\end{figure}

After meticulous filtering, our dataset contains {99 scenes}, {2,014 episodes} derived from {1290 unique question-answer pairs}.
Fig.~\ref{fig:dataset_statistics} shows that the dataset covers diverse scene categories and exhibits a wide distribution of question and answer lengths.
The final benchmark focuses on six question categories that broadly evaluate an agent’s capability in active perception and spatial reasoning, illustrated in Fig.~\ref{fig:dataset_samples}.
These categories test an agent's ability to (1) perform Object Search (OS) by locating items explicitly mentioned in the question, (2) recognize Object States (OST) such as open or closed, which requires navigating to a viewpoint where the state is observable, (3) identify Object Attributes (OA) such as physical characteristics and features, (4) conduct Context-guided Search (CGS) to find objects by functional or situational context when the object name is not given, (5) perform Spatial Reasoning (SR) over relative positions and sizes, and (6) perform Counting (CNT) over instances of an object class.

In addition, the action distribution in Fig.~\ref{fig:dataset_statistics}(c) shows that all action types are well utilized, including those beyond planar navigation, suggesting the need for 5-DoF active viewpoint control.
Notably, \textit{move\_backward} is more frequent than \textit{move\_forward}, indicating that gaining context by moving away from the target is often necessary.
Moreover, \textit{move\_up} is the most frequent action, and \textit{look\_down} appears more often than \textit{look\_up}, suggesting that relevant visual evidence is often obtained by observing target objects from lateral or slightly elevated viewpoints rather than from below.

\section{Evaluation Protocol and Baselines}
We design evaluation settings to assess whether VLMs can actively acquire visual evidence through viewpoint control, resolving spatial ambiguities such as occlusions and depth uncertainty by selecting informative viewpoints.

\subsection{E3VS Framework for VLMs}
\label{subsec:proposed_method}
To evaluate VLMs on the E3VS task, we implement a closed-loop perception–action framework in which the model iteratively selects actions based on egocentric observations.
At each time step $t$, the agent receives an observation $O_t$ rendered from the 3DGS scene $G$ at its viewpoint $v_t = (x, y, z, \theta, \phi)$, and the VLM processes $O_t$ with the question to select an action $a_t \in \mathcal{A}$. This continues until the agent executes \texttt{stop}, after which it generates the final answer.

To interface VLMs with the environment, we construct a structured prompt that conditions the model on both the environment configuration and the current interaction state.
The prompt consists of two components:
\begin{itemize}
    \item \textbf{System Prompt:} Defines the world coordinate system (Z-axis as up) and the constraints of the action space, including fixed translation ($0.25\,\mathrm{m}$) and rotation ($30^\circ$) intervals.

    \item \textbf{User Prompt:} Provides task input and dynamic state information, including the question, the current observation image, step count, 3D coordinates, the previous action, and feedback such as collision detection.
\end{itemize}
Unless otherwise specified, our experiments use a single-frame observation setting without additional reasoning modules to establish a clear baseline.
Full prompt templates are provided in the Appendix.

\subsection{Baseline Models} 
\label{subsec:baseline_setting}
To examine different levels of perceptual access and spatial reasoning, we define five categories of baseline agents:
(1) blind VLMs, (2) VQA at Start, (3) VQA at Birdview, (4) VQA at Goal, and (5) 2D Visual Search at Start.
These baselines form a progressive spectrum of perceptual conditions, from no visual input to privileged viewpoints, providing a reference for interpreting embodied agent performance on E3VS-Bench.

\vspace{0.1cm}
\noindent \textbf{(1) Blind VLMs.}
In this setting, models are asked to answer questions using only textual input without any visual observation.
This baseline measures how well models can infer answers from linguistic cues and prior knowledge alone.

\vspace{0.1cm}
\noindent \textbf{(2--4) VQA from Fixed Viewpoints.}
In these settings, VLMs perform visual question answering from a single static image captured at a predefined viewpoint, without any viewpoint movement.

\textbf{VQA at Start} uses the image from the initial viewpoint and evaluates whether the question can be answered without exploration.

\textbf{VQA at Goal} uses an image from human-annotated answerable viewpoints, representing an upper bound where sufficient visual evidence is available.

\textbf{VQA at Birdview} provides a bird's-eye image from directly above the scene, evaluating whether a global overview alone suffices. Many questions in E3VS-Bench still require fine-grained, viewpoint-dependent evidence (e.g., text, object states, occluded regions) that cannot be resolved without moving to appropriate viewpoints.

\vspace{0.1cm}
\noindent \textbf{(5) 2D Visual Search at Start.}
This baseline examines whether questions can be solved by image-space exploration alone, without 3D viewpoint transitions. We adopt representative methods such as SEAL~\cite{Wu2024vstar} and DyFo~\cite{dyfo2025} to show the limitations of image-based search relative to embodied 3D visual search.

\section{Experiments}

\subsection{Experimental Setting}


\vspace{0.1cm}
\noindent \textbf{Dataset Split.}
The dataset is split into train, validation, and test sets at the scene level.
The train set contains 1,406 episodes from 900 question–answer pairs across 68 scenes, the validation set contains 231 episodes from 132 pairs across 10 scenes for model selection and hyperparameter tuning, and the test set contains 377 episodes from 258 pairs across 21 scenes for final evaluation.
All splits are going to be publicly released.
In this work, we focus on evaluating zero-shot VLM-based frameworks.

\vspace{0.1cm}
\noindent \textbf{Evaluation Metrics.}
We evaluate models from three aspects: answer correctness, exploration efficiency and navigation safety.
Regarding answer correctness, we employ a VLM-as-a-judge framework in accordance with OpenEQA~\cite{mahumar2024openeqa}, using \textit{GPT 5.1} as the evaluator.
The judge VLM receives the predicted response and ground-truth answer, along with the end and goal images, and outputs a score of 5 for correct predictions and 1 for incorrect ones.
This metric shows a Spearman correlation of $\rho=0.54$ with human evaluation, indicating reasonable agreement.
Exploration efficiency is quantified by the average number of steps, and navigation safety by Collision Rate, a binary indicator that is 1 if any collision occurs within an episode and 0 otherwise.

\vspace{0.1cm}
\noindent \textbf{Baseline Settings.}
Across the baseline settings, we evaluate a diverse set of proprietary and open-source models, 
including \textit{Gemini 2.5 Pro/Flash}, \textit{Gemini 3.0 Pro/Flash}, \textit{GPT 5.1}, 
and open-source models such as \textit{Qwen3-VL 8B/30B}, \textit{InternVL3.5-8B}, and \textit{Step3-VL-10B}. 
These models are evaluated under embodied E3VS agent settings, 
while a subset of models is used for static baselines to enable controlled comparison.
For static baseline settings, we evaluate representative models including 
\textit{Gemini 2.5 Flash}, \textit{GPT 5.1}, and \textit{Qwen3-VL-8B}. 
To avoid evaluation bias, \textit{GPT 5.1} is excluded from the blind and start-view settings, since it was used during dataset filtering. For the same reason, \textit{GPT 5.1} is used for answer generation in all E3VS agent conditions except Human: because episodes it could answer from the start viewpoint were already removed (Section~\ref{sec:data_construction_pipeline}), using a different generator would introduce model-specific bias.

\vspace{0.1cm}
\noindent \textbf{Implementation Details.}
We use images with a resolution of $512 \times 512$ pixels and a field of view (FOV) of $90^\circ$.
Each episode is limited to a maximum of 25 steps, providing a sufficient exploration budget while preventing excessively long trajectories.
For locomotion actions, the agent moves $0.25\,\mathrm{m}$ per step, while rotation actions rotate the camera by $30^\circ$ in place.
The maximum output token length is 128 when reasoning is disabled and 256 or more when enabled, depending on the model. If no valid action is produced within this limit, the agent defaults to \texttt{move\_forward}.

\subsection{Main Results}

\begin{table}[t]
\centering
\renewcommand{\arraystretch}{0.95}
\begin{minipage}{\columnwidth} 
\setlength{\tabcolsep}{1.2pt}
\footnotesize
\begin{tabularx}{\columnwidth}{l *{9}{>{\centering\arraybackslash}X}}
\toprule
\textbf{Model} & \textbf{OS} & \textbf{OST} & \textbf{OA} & \textbf{CGS} & \textbf{SR} & \textbf{CNT} & \textbf{Avg.} & \textbf{Steps} & \textbf{Coll.} \\
\midrule
Random Action & 2.38 & 2.18 & 2.60 & 2.20 & 2.50 & 2.00 & 2.28 & 10.15 & 0.39 \\ 
\midrule
\textit{Proprietary Models} & & & & & & & &  &  \\
Gemini 2.5 Pro & 3.14 & 2.36 & 2.45 & 2.60 & 2.25 & \textbf{2.04} & 2.54 & 7.80 & 0.31 \\
Gemini 3.0 Flash & \textbf{3.21} & \textbf{2.82} & \textbf{3.18} & \textbf{3.53} & 2.75 & 1.88 & \textbf{2.79} & 11.29 & 0.43 \\
Gemini 3.0 Pro & 3.07 & 2.55 & \textbf{3.18} & 3.40 & \textbf{3.00} & 1.96 & 2.75 & 10.17 & 0.34 \\
GPT 5.1 & 2.90 & 2.18 & 2.53 & 2.73 & 2.25 & 1.88 & 2.42 & 15.04 & 0.49 \\
\midrule
\textit{Open Source Models} & & & & & & & &  &  \\
Qwen3-VL-8B~\cite{Qwen3-VL} & 2.86 & 2.36 & 2.60 & 3.13 & 1.88 & 1.96 & 2.46 & 18.73 & 0.30 \\
Qwen3-VL-30B~\cite{Qwen3-VL} & 2.67 & 2.09 & 2.60 & 2.47 & 2.25 & 1.84 & 2.32 & 14.18 & 0.50 \\
InternVL3.5-8B~\cite{wang2025internvl35} & 2.66 & 2.18 & 2.24 & 2.60 & 2.12 & 1.60 & 2.21 & 10.85 & 0.65 \\
Step3-VL-10B~\cite{huang2026step3vl10btechnicalreport} & 2.38 & 2.27 & 2.38 & 3.13 & 2.50 & 1.64 & 2.24 & 11.16 & 0.42 \\
\midrule
Human & 3.12 & 3.59 & 4.06 & 4.06 & 3.35 & 3.59 & 3.53 & 11.21 & - \\
\bottomrule
\end{tabularx}
\end{minipage}
\caption{
Main results on E3VS-Bench. 
We evaluate various baseline models across six fine-grained question types.
All values except Steps and Coll. are VLM judge scores on a scale of 1 to 5 where a higher score indicates better performance.
Steps and Coll. denote the average number of Navigation Steps and collision rate respectively.
}
\label{tab:3dvs_results}
\vspace{-2.25em}
\end{table}

We evaluate a wide range of VLMs on the E3VS-Bench, categorized into static baselines and E3VS agents. The results are summarized in Table~\ref{tab:3dvs_results},~\ref{tab:vqa_results}.

\vspace{0.1cm}
\noindent \textbf{Performance Across Model Families on E3VS.}
The evaluation on E3VS shows that \textit{Gemini 3.0 Flash} substantially outperforms all other models across nearly all categories.
In contrast, \textit{GPT 5.1} performs comparably to open-source models such as \textit{Qwen3-VL-8B} and \textit{Step3-VL-10B}, and overall these models achieve scores only slightly above a Random Action baseline.
This suggests that while current VLMs possess strong 2D image recognition capabilities, they show limitations in 3D visual search settings that require viewpoint changes.
A substantial gap also remains between current models and human performance.

\begin{table}[t]
\centering
\footnotesize
\renewcommand{\arraystretch}{0.95}
\resizebox{\columnwidth}{!}{
\setlength{\tabcolsep}{3pt} 
\begin{tabular*}{\columnwidth}{@{\extracolsep{\fill}} l *{6}{wc{1.0cm}} p{1.2cm}}
\toprule
\textbf{Model} & \textbf{OS} & \textbf{OST} & \textbf{OA} & \textbf{CGS} & \textbf{SR} & \textbf{CNT} & \textbf{Avg.} \\
\specialrule{.1em}{.05em}{.05em} 
\rowcolor{gray!10} \multicolumn{8}{l}{\textit{Static Baselines (No Agent Movement)}} \\
\midrule
\rowcolor{gray!5} \multicolumn{8}{l}{\textbf{Blind VLM}} \\
Gemini 2.5 Flash & 1.72 & 2.00 & 2.09 & 1.67 & 1.88 & 1.72 & 1.82 \\
Qwen3-VL-8B & 1.38 & 2.64 & 1.58 & 2.20 & 1.50 & 1.52 & 1.67
 \\
\rowcolor{gray!5} \multicolumn{8}{l}{\textbf{2D Visual Search at Start}} \\
SEAL~\cite{Wu2024vstar} & 1.62 & 2.45 & 1.44 & 1.53 & 2.50 & 1.32 & 1.68 \\
DyFO~\cite{dyfo2025} & 1.90 & 2.45 & 1.44 & 1.67 & 2.00 & 1.80 & 1.86 \\
\rowcolor{gray!5} \multicolumn{8}{l}{\textbf{VQA at Start}} \\
Gemini 2.5 Flash & 2.21 & 2.64 & 1.73 & 2.33 & 2.12 & 2.00 & 2.14 \\
Qwen3-VL-8B & 2.10 & 2.82 & 1.87 & 2.20 & 2.12 & 1.56 & 2.02 \\
\rowcolor{gray!5} \multicolumn{8}{l}{\textbf{VQA at Birdview}} \\
Gemini 2.5 Flash & 1.31 & 1.64 & 1.44 & 1.67 & 1.75 & 1.48 & 1.48\\
GPT 5.1 & 1.48 & 2.09 & 1.58 & 2.20 & 1.62 & 1.16 & 1.55 \\
Qwen3-VL-8B & 1.55 & 2.18 & 2.09 & 1.80 & 1.38 & 1.12 & 1.59 \\
\rowcolor{gray!5} \multicolumn{8}{l}{\textbf{VQA at Goal}} \\
Gemini 2.5 Flash & 3.79 & 3.27 & 3.84 & 4.20 & 3.50 & 3.16 & 3.58 \\
GPT 5.1 & 4.17 & 3.36 & 3.91 & 4.20 & 3.00 & 2.88 & 3.60 \\
Qwen3-VL-8B & 3.07 & 3.55 & 3.40 & 4.20 & 3.00 & 2.20 & 3.03 \\
Human & 4.25 & 3.82 & 4.47 & 4.25 & 4.14 & 3.82 & 4.12 \\
\bottomrule
\end{tabular*}
}
\caption{
Comparison of static baselines on E3VS-Bench. 
The large performance gap between `VQA at Start' and `VQA at Goal' indicates that initial viewpoints are often insufficient, necessitating active exploration.
}
\label{tab:vqa_results}
\vspace{-2.4em}
\end{table}

\vspace{0.1cm}
\noindent \textbf{Comparison of OS and CGS.}
Models often perform better on CGS than on explicit OS (Table~\ref{tab:3dvs_results}).
CGS queries can often be solved by recognizing relevant objects or functional zones (e.g., ``where to store binders'').
In contrast, OS requires not only recognizing the target object but also understanding its spatial relations and placement in the scene.
This additional spatial reasoning likely increases task complexity and leads to lower performance.

\vspace{0.1cm}
\noindent \textbf{Challenges in CNT.}
In CNT, models already struggle in the 2D single-view setting and degrade further in the full E3VS setting (Tables~\ref{tab:vqa_results},~\ref{tab:3dvs_results}), whereas humans score 3.82 given the goal viewpoint. This reveals a fundamental limitation: counting requires accumulating evidence across a 5-DoF trajectory, which single-observation EQA approaches cannot handle.

\vspace{0.1cm}
\noindent \textbf{Emerging Viewpoint Capability for SR.}
SR questions require understanding geometric relationships such as relative positions, distances, and size comparisons between objects, and the quality of such geometric evidence is highly viewpoint-dependent; comparing object heights, for example, is far easier from a lateral viewpoint than from an oblique or top-down view. In this context, \textit{Gemini 3.0 Pro} achieves E3VS performance (Table~\ref{tab:3dvs_results}) comparable to the VQA at Goal upper bound (Table~\ref{tab:vqa_results}), indicating that it can not only recognize spatial relationships but also actively select viewpoints that improve their observability, effectively integrating viewpoint selection into SR.

\vspace{0.1cm}
\noindent \textbf{Limited Gain from Viewpoint Selection in OST.}
OST questions typically involve binary decisions (e.g., open vs. closed), which can often be predicted from prior knowledge or dataset bias even without sufficient visual evidence. 
Importantly, during dataset construction, episodes that \textit{GPT 5.1} could answer correctly from the initial viewpoint were filtered out.
However, similar filtering was not applied to other models such as Qwen or Gemini.
As a result, VQA at Start performance for these models still reflects the influence of bias.
Empirically, we observe that the VQA at Start performance of \textit{Qwen3-VL-8B} is comparable to the E3VS performance of \textit{Gemini 3.0 Flash} evaluated with \textit{GPT 5.1} as the judge.
This suggests that active viewpoint selection yields only marginal improvements beyond compensating for such biases.
Furthermore, the viewpoint required to resolve the state is relatively predictable (e.g., Fig.~\ref{fig:dataset_samples}), reducing OST to viewpoint selection and path planning rather than exploration. Despite this, gains remain limited, indicating that models still struggle to reach the appropriate viewpoint even when it is relatively clear.

\vspace{0.1cm}
\noindent \textbf{High Viewpoint Sensitivity in OA.}
OA questions require identifying fine-grained, instance-specific properties of a target object, such as text, material, color, or subtle visual features.
While the target is explicitly specified (as in OS), it is often unclear from the initial viewpoint where the relevant evidence resides on the object. Unlike OST, whose observation region is predictable (e.g., a door for open/closed), OA thus requires actively exploring the object to discover informative evidence for each instance.
Empirically, while \textit{Gemini 3.0} models achieve relatively high performance on OA, other models perform at or below the Random Action baseline.
This highlights the difficulty of acquiring fine-grained visual evidence through 5-DoF active viewpoint control, suggesting that OA is highly sensitive to viewpoint selection and requires both exploratory viewpoint discovery and detailed visual recognition.

\subsection{Ablation Study}
\label{sec:ablation}
\begin{table}[t]
\centering
\renewcommand{\arraystretch}{0.95}
\begin{minipage}{\columnwidth} 
\setlength{\tabcolsep}{0.9pt}
\footnotesize
\begin{tabularx}{\columnwidth}{l *{9}{>{\centering\arraybackslash}X}}
\toprule
\textbf{Model} & \textbf{OS} & \textbf{OST} & \textbf{OA} & \textbf{CGS} & \textbf{SR} & \textbf{CNT} & \textbf{Avg.} & \textbf{Steps} & \textbf{Coll.} \\
\midrule
Gemini 3.0 Flash & 3.21 & 2.82 & 3.18 & 3.53 & 2.75 & 1.88 & 2.79 & 11.29 & 0.43 \\
Gemini 3.0 Flash+Th. & 3.21 & 2.82 & 2.96 & 3.27 & 3.00 & 2.00 & 2.79 & 19.80 & 0.41\\
GPT 5.1 & 2.90 & 2.18 & 2.53 & 2.73 & 2.25 & 1.88 & 2.42 & 15.04 & 0.49 \\
GPT 5.1+Th. & 2.97 & 2.64 & 3.25 & 2.87 & 2.38 & 2.16 & 2.70 & 13.78 & 0.38 \\
\bottomrule
\end{tabularx}
\end{minipage}
\caption{
Ablation study on the effect of the thinking process on E3VS-Bench.
Th. indicates whether the model performs the Thinking process.
}
\label{tab:3dvs_results_thinking_comparison}
\vspace{-1.25em}
\end{table}

\noindent \textbf{Impact of Internal Reasoning.}
We analyze the effect of thinking-related parameters on E3VS performance without using explicit reasoning prompts (Table~\ref{tab:3dvs_results_thinking_comparison}).
For \textit{GPT 5.1}, increasing the thinking budget consistently improves performance, suggesting that additional internal computation benefits viewpoint planning and decision making.
In contrast, \textit{Gemini 3.0 Flash} shows little sensitivity to the reasoning budget.
Overall, the effectiveness of increased reasoning appears to be model-dependent, though the cause remains unclear and may relate to differences in architecture or training data.

\begin{figure}[t]
\centering
\includegraphics[width=\columnwidth]{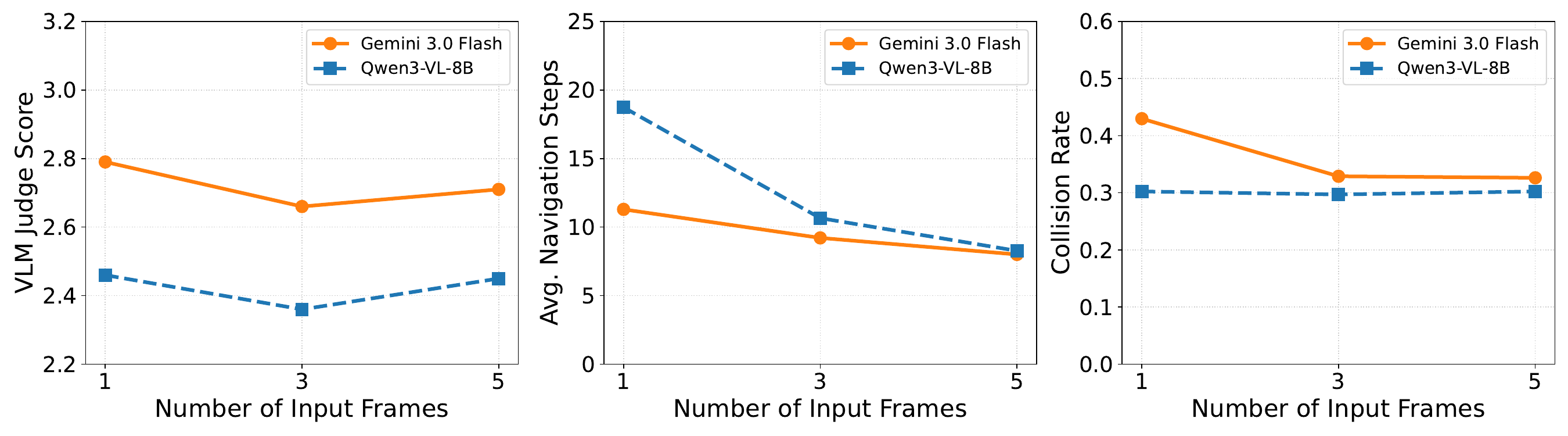}
\vspace{-1.5em}
\caption{
Effect of the number of input frames on E3VS performance.
While more frames do not significantly impact the VLM judge score, they consistently lead to more efficient navigation (fewer steps) and safer trajectories (lower collision rates).
}
\label{fig:ablation_multi_frames}
\vspace{-1.25em}
\end{figure}

\vspace{0.1cm}
\noindent \textbf{Effect of Memory.}
We study the effect of temporal information by varying the number of input frames (1, 3, and 5) for \textit{Gemini 3.0 Flash} and \textit{Qwen3-VL-8B}.
As shown in Fig.~\ref{fig:ablation_multi_frames}, multi-frame inputs do not significantly improve VLM Judge scores, but consistently reduce navigation steps and collision rates.
Step reduction is likely due to fewer action ``deadlocks,'' where single-frame agents repeat the same actions.
Collision reduction appears to result from improved action–observation understanding, enabling better anticipation of viewpoint changes and more effective collision avoidance.

\begin{table}[t]
\centering
\renewcommand{\arraystretch}{0.95}
\setlength{\tabcolsep}{2.5pt} 
\footnotesize

\begin{tabularx}{\columnwidth}{l c *{7}{>{\centering\arraybackslash}X} c c}
\toprule
\textbf{Model} & \textbf{Init. at Goal} & \textbf{OS} & \textbf{OST} & \textbf{OA} & \textbf{CGS} & \textbf{SR} & \textbf{CNT} & \textbf{Avg.} & \textbf{Steps} & \textbf{Coll.} \\
\midrule
Qwen3-VL-8B & \xmark & 2.86 & 2.36 & 2.60 & 3.13 & 1.88 & 1.96 & 2.46 & 18.73 & 0.30 \\
Qwen3-VL-8B & \checkmark & 3.90 & 3.36 & 3.84 & 3.93 & 2.75 & 2.56 & 3.38 & 12.06 & 0.23 \\
Gemini 3.0 Flash & \xmark & 3.21 & 2.82 & 3.18 & 3.53 & 2.75 & 1.88 & 2.79 & 11.29 & 0.43 \\
Gemini 3.0 Flash & \checkmark & 4.21 & 3.55 & 3.33 & 3.93 & 3.25 & 2.38 & 3.41 & 6.60 & 0.29 \\
\bottomrule
\end{tabularx}
\caption{
Ablation study with goal-initialized viewpoints, where the initial viewpoint is set to the goal viewpoint containing sufficient visual evidence. This removes the need for exploration and isolates the model's ability to recognize visual evidence and make appropriate stopping decisions.
}
\label{tab:3dvs_results_start_from_answerable}
\vspace{-2.4em}
\end{table}

\vspace{0.1cm}
\noindent \textbf{Decoupling Recognition and Exploration via Observable Evidence.}
We investigate whether VLMs can correctly recognize visual evidence and terminate the episode when the required information is already visible from the current viewpoint.
To this end, we construct a setting where the initial view contains sufficient visual evidence to answer the question, removing the need for further exploration.
As shown in Table~\ref{tab:3dvs_results_start_from_answerable}, both models achieve strong performance under this condition, approaching the VQA at Goal performance of \textit{GPT 5.1} reported in Table~\ref{tab:vqa_results}.
However, their stopping behaviors differ notably.
\textit{Gemini 3.0 Flash} terminates efficiently, requiring only 6.60 steps on average, whereas \textit{Qwen3-VL-8B} takes nearly twice as many steps (12.06), suggesting that \textit{Qwen3-VL-8B} tends to continue exploring even when sufficient visual evidence is already available.
Across most question types, \textit{Gemini 3.0 Flash} outperforms \textit{Qwen3-VL-8B}, indicating stronger overall evidence utilization.
In contrast, for OA questions, \textit{Qwen3-VL-8B} achieves substantially higher scores, implying a stronger ability to recognize fine-grained object properties and make stopping decisions based on such cues.
Interestingly, this trend reverses in the standard E3VS setting with unanswerable initial viewpoints, where Gemini consistently outperforms \textit{Qwen3-VL-8B}.
This suggests that the two models exhibit different strengths:
Gemini is more effective at actively exploring to acquire missing information,
whereas \textit{Qwen3-VL-8B} shows stronger capability in recognizing fine-grained visual attributes when evidence is present, but does not consistently translate this recognition into appropriate stopping behavior.

\subsection{Qualitative Results}

\begin{figure}[t]
\centering
\includegraphics[width=\columnwidth]{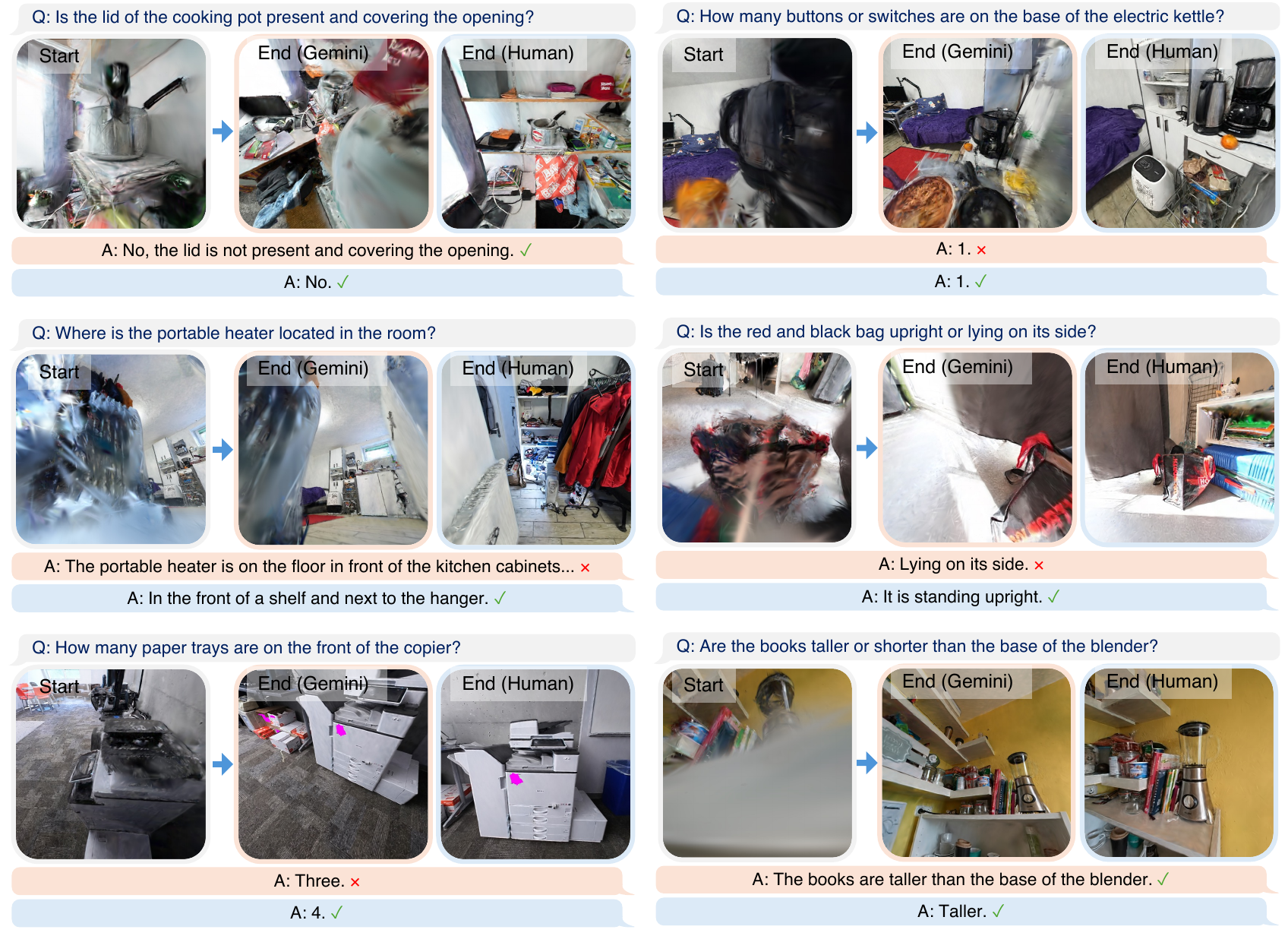}
\vspace{-1.5em}
\caption{
Qualitative Results.
The orange bars represent the predicted answers after visual search by \textit{Gemini 3.0 Flash}, while the blue bars indicate the human performance.
}
\vspace{-1.25em}
\label{fig:qualitative_results}
\end{figure}

Fig.~\ref{fig:qualitative_results} compares the viewpoints selected by humans and \textit{Gemini 3.0 Flash}.
Human-selected viewpoints are typically intuitive, deliberately centering the target object or revealing the discriminative features required to answer the question.
In contrast, the ``answerable'' viewpoints reached by agents are frequently sub-optimal, as they do not always capture sufficient visual evidence even when the target object is partially visible.
Since the VLM-as-a-judge penalizes answers whose supporting evidence is absent from the final observation, such stops receive low scores (e.g., Fig.~\ref{fig:qualitative_results}, first row, second column, where \textit{Gemini 3.0 Flash} halts before the task-relevant features are visible). This highlights the importance of actively selecting evidence-revealing viewpoints, which remains challenging for current VLM-based agents.
\section{Conclusion}
\label{6_conclusion}

In this paper, we introduced \textbf{E3VS-Bench}, a benchmark for viewpoint-dependent active perception in photorealistic 3D Gaussian Splatting environments. 
E3VS requires agents to manipulate viewpoints in 5-DoF to acquire task-critical evidence.
The benchmark consists of 99 high-fidelity scenes and 2,014 episodes covering diverse reasoning types, including object search, spatial reasoning, attribute recognition, and counting.
We evaluated a range of state-of-the-art proprietary and open-source VLMs under both static and active settings. 
While several models demonstrate strong 2D visual reasoning ability, all evaluated systems exhibit a substantial performance gap compared to human performance on E3VS. 
These results suggest that current VLM-based agents still lack robust viewpoint planning in 3D environments.

\noindent \textbf{Limitations.}
We adopt a VLM-as-a-judge framework to evaluate open-vocabulary answers. 
However, we observe a relatively weak correlation between human evaluation and VLM-based judging. 
In some cases, the agent’s final observation differs substantially from the human-annotated goal viewpoint while receiving a similar judge score, indicating that current automated evaluation may insufficiently capture viewpoint adequacy.

\noindent \textbf{Future Work.}
This work focuses on zero-shot evaluation without exploiting the train/validation splits for optimization. 
We release each split to facilitate future learning-based agents.

\noindent \textbf{Acknowledgments.}
This work was supported by JST PRESTO (Grant Number JPMJPR22P8), JST CRONOS (Grant Number JPMJCS24K6), JST SPRING (Grant Number JPMJSP2108), JSPS KAKENHI (Grant Number 25K03177), and JST K Program (Grant Number JPMJKP25V2), Japan.


\bibliographystyle{splncs04}
\bibliography{main}
\newpage

\setcounter{page}{1}
\appendix

\section{Appendix}

This supplementary material provides additional details that complement the main paper.
We describe the dataset construction pipeline, implementation details for E3VS,
the prompts used in the E3VS framework, and the evaluation protocol used for VLM-as-a-judge.

\section{Dataset Construction Pipeline Details}

The overall dataset construction pipeline is described in the main paper. 
In this section, we provide additional details for the stages that involve automated filtering or manual verification, including viewpoint filtering, QA generation, human annotation, initial viewpoint filtering, and counting modification.






\subsection{Viewpoint Filtering and QA Generation}

\begin{figure}[h]
\centering
\includegraphics[width=0.9\linewidth]{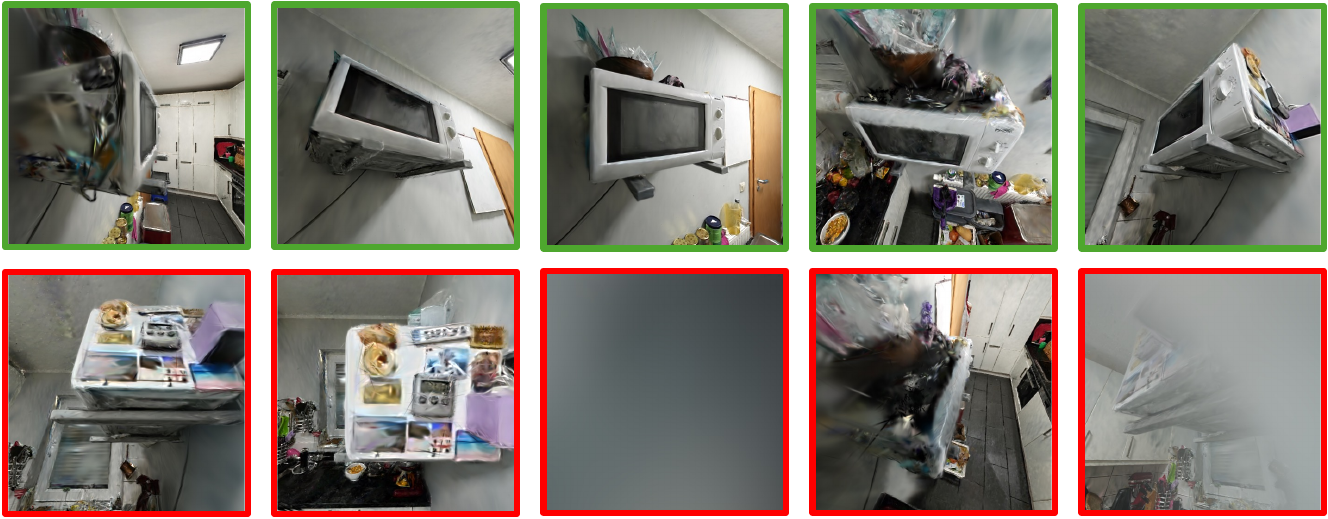}
\caption{
Qualitative examples of viewpoint filtering by the VLM.
Accepted viewpoints are highlighted with green bounding boxes, while filtered viewpoints are highlighted with red bounding boxes.
}
\label{fig:prompt_viewpoint_filtering}
\end{figure}

In Step 2 of the dataset construction pipeline, we filter viewpoints for QA generation following 3D scene curation (Step 1). 
We retain viewpoints that satisfy the \textit{same\_object} condition, which requires that the target object class is clearly visible in the image rather than ambiguous. 
Viewpoints that do not satisfy this condition are discarded. 
This criterion also implicitly removes many invalid viewpoints, such as those outside the scene or penetrating scene geometry, where the target object cannot be clearly observed. 
The prompt used for this stage is shown in Fig.~\ref{fig:target_object_visibility_detection_prompt} and qualitative examples of the viewpoint filtering results are shown in Fig.~\ref{fig:prompt_viewpoint_filtering}.

\begin{figure*}[t]
\centering
\begin{minipage}{0.97\linewidth}

\begin{mdframed}[
  backgroundcolor=blue!10,
  linecolor=red!10,
  linewidth=0pt,
  innerleftmargin=10pt,
  innerrightmargin=10pt,
  innertopmargin=6pt,
  innerbottommargin=6pt,
  skipabove=0pt,
  skipbelow=0pt
]
\centering
\bfseries
Prompt for Target Object Visibility Detection.
\end{mdframed}

\begin{mdframed}[
  backgroundcolor=blue!2,
  linecolor=white,
  linewidth=0pt,
  innerleftmargin=14pt,
  innerrightmargin=14pt,
  innertopmargin=10pt,
  innerbottommargin=10pt,
  skipabove=0pt,
  skipbelow=0pt
]
\footnotesize

You are an expert in object detection. Please analyze this image and determine if a [OBJECT\_CLASS\_NAME] is clearly visible in the image.

Please answer the following question with only 'Yes' or 'No'.
Is there a [OBJECT\_CLASS\_NAME] clearly visible in this image?

Important:
\begin{itemize}
    \item Answer ``Yes'' only if you can clearly see a [OBJECT\_CLASS\_NAME] in the image
    \item Answer ``No'' if the [OBJECT\_CLASS\_NAME] is not visible, partially visible, or unclear
    \item Ignore image quality, lighting, or rendering artifacts - focus only on object visibility.
\end{itemize}

Your answer must be only ``Yes'' or ``No''.

A second image is provided as a binary mask (white=target, black=background). Use it as additional context.

\end{mdframed}
\end{minipage}
\caption{Prompt for Target Object Visibility Detection.}
\label{fig:target_object_visibility_detection_prompt}
\end{figure*}

\begin{figure}[h]
\centering
\includegraphics[width=\linewidth]{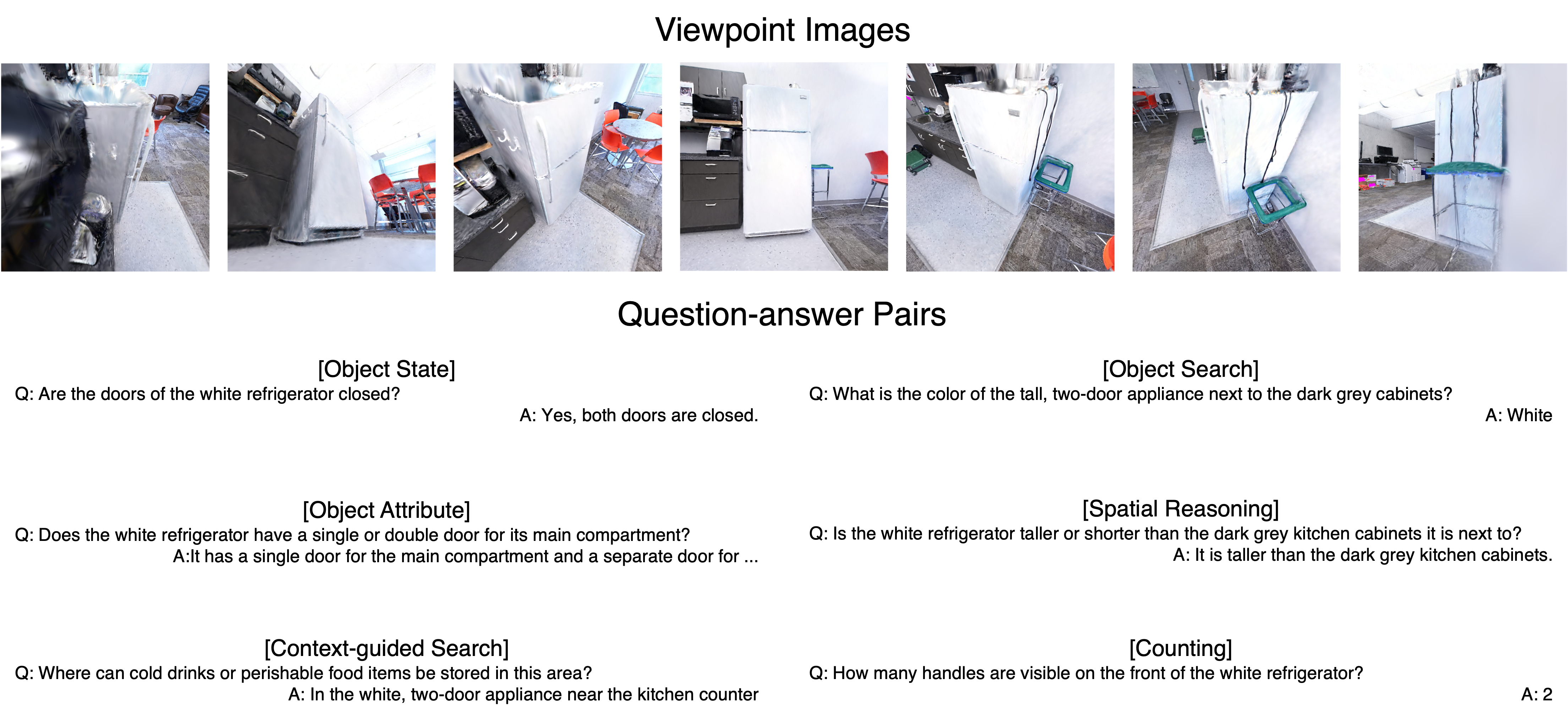}
\caption{
Examples of Generated Question-answer pairs by \textit{Gemini 2.5 Flash}.
}
\label{fig:samples_qa_generation}
\end{figure}

The filtered images are then provided to the VLM to generate QA candidates. 
The prompt used for QA generation is shown in Fig.~\ref{fig:prompt_qa_generation}, and examples of generated QA pairs are presented in Fig.~\ref{fig:samples_qa_generation}.

\begin{figure*}[t]
\centering
\begin{minipage}{0.97\linewidth}

\begin{mdframed}[
  backgroundcolor=blue!10,
  linecolor=red!10,
  linewidth=0pt,
  innerleftmargin=10pt,
  innerrightmargin=10pt,
  innertopmargin=6pt,
  innerbottommargin=6pt,
  skipabove=0pt,
  skipbelow=0pt
]
\centering
\bfseries
Prompt for Generating Subset Sensitive QAs.
\end{mdframed}

\begin{mdframed}[
  backgroundcolor=blue!2,
  linecolor=white,
  linewidth=0pt,
  innerleftmargin=14pt,
  innerrightmargin=14pt,
  innertopmargin=10pt,
  innerbottommargin=10pt,
  skipabove=0pt,
  skipbelow=0pt
]
\scriptsize

Reference images A1..A[NUM\_OF\_IMAGES] follow:
Question category descriptions:
\begin{itemize}
    \item \textit{Object Search}: Questions that locate the target object or ask about its position (e.g., 'What is under the table?' 'Where is the red box?').
    \item \textit{Context-guided Search}: Questions that infer the target object from functional context without naming it explicitly (e.g., 'Where can I wash my hands?' 'Where could I place dirty dishes?').
    \item \textit{Object Attribute}: Questions about observable attributes of the object such as color, shape, texture, labels, or count (e.g., 'What color is the lamp?' 'Does the keyboard have a numeric pad?').
    \item \textit{Object State}: Questions about the state or condition of the object, including open/closed, on/off, running/stopped, cluttered/organized, etc. (e.g., 'Is the door closed?' 'Is the water running?').
    \item \textit{Spatial Reasoning}: Questions about 3D spatial relationships, relative positions, or comparisons between objects (e.g., 'Which is taller, the speaker or the plant?').
    \item \textit{Counting}: Questions about the number of objects (e.g., 'How many balls are there on the large wooden desk?').
\end{itemize}

Object list in this scene: [NUM\_OF\_THE\_INSTANCES] [INSTANCE\_CATEGORY], ...
 (use this list specifically for Context-guided Search).

You will see [NUM\_OF\_IMAGES] images of the same object instance (A1..A[NUM\_OF\_IMAGES]).
Propose creative and diverse questions about the [OBJECT\_CATEOGY] only.
The goal is to generate a wide variety of questions, covering different aspects of the object.
Think about the object's function, its parts, its relationship to its environment, and any interesting details you can observe. "
Questions must be strictly about properties or parts of the target object itself, not the background scene, lighting, camera, reflections, shadows, or other objects.
IMPORTANT: Formulate questions that help identify and distinguish this specific target object within a scene.
For Context-guided Search questions, focus on the functional use or purpose without explicitly naming the object type.
For other question types, include identifying attributes such as: color, size, shape, material, brand, text/labels, specific features, or unique characteristics,
OR include spatial or positional context such as: orientation, placement, position relative to other objects, or spatial relationships.
For example, for Context-guided Search: 'Where would be the best place to prepare food?' instead of 'What color is the kitchen counter?'.
For other types: 'What color is the cylindrical bottle on the table?' or 'What brand name is written on the red object?'.
Propose questions that can be answered using only a subset of these images (i.e., answerable for some images but not all).
Prefer questions relying on view-dependent details (e.g., small logos, text, backside color, connection points, ports, visible buttons).
DO NOT create questions about damage, scratches, wear, defects, or any quality issues that might be caused by 3D reconstruction artifacts.
For each question, provide the correct answer based on what you can observe in the reference images of the [OBJECT\_CATEOGY].
The answer should be concise and factual, describing what is visible in the images where the question is answerable.

Generate exactly one question for EACH of the following categories:
\textit{Object Search}, \textit{Context-guided Search}, \textit{Object Attribute}, \textit{Object State}, \textit{Spatial Reasoning}, \textit{Counting}.
Return a strict JSON array 'questions' with up to 6 items. Each item must include keys: category (one of the listed), question (string), answer (string). No extra text.

\end{mdframed}
\end{minipage}
\caption{Prompt for Generating Subset Sensitive QAs.}
\label{fig:prompt_qa_generation}
\end{figure*}

\subsection{Invalid QA Filtering}

In Step 3 of the dataset construction pipeline, human annotators verify automatically generated question--answer (QA) pairs and remove invalid questions.  
For the remaining valid questions, annotators also correct the associated answers when necessary.

\paragraph{Annotation Setup.}
For each QA candidate, annotators are provided with 
(1) a video capturing the entire scene,
(2) the question and its pre-generated answer,
(3) candidate viewpoint images around the target object, and
(4) the object category of the target instance.  

\paragraph{Invalid QA Filtering.}
A question is considered valid only if all of the following conditions are satisfied; otherwise the QA pair is removed.

\begin{enumerate}

\item \textbf{Single-View Answerability.}  
The question must be answerable from a single image observation.  
Questions requiring multiple viewpoints or lacking sufficient visual evidence are rejected.

\item \textbf{Question Validity.}  
The description in the question must correctly reflect the attributes or state of the target object.  
Questions containing incorrect descriptions are rejected.

\item \textbf{Object Specificity.}  
The question must uniquely identify the target object within the scene.  
Questions are rejected if multiple similar objects exist and the target cannot be unambiguously determined.

\item \textbf{Object Centering.}  
The target object should appear near the center of the image.  
Questions are rejected if the object is outside the frame or located at an extreme boundary.

\item \textbf{Instance Name Matching.}  
The object referenced in the question must correspond to the annotated instance name of the target object.

\item \textbf{Image ID Exclusion.}  
Questions must not explicitly refer to image identifiers such as ``A1'' or ``A2''.  
For example: \textit{``What is the color of the rectangular frame surrounding the window glass in images A1 and A2?''}

\end{enumerate}

\paragraph{Answer Correction.}
After invalid questions are removed, annotators review the pre-generated answers for the remaining QA pairs.
If an answer is incorrect, it is revised to reflect the correct observation from the images.
Answers are required to be concise, objective, and based only on visually observable information.
Speculative or inferred descriptions that are not directly supported by visual evidence are not allowed.


\subsection{Viewpoint Labeling}
\label{sec:appendix_viewpoint_labeling}

After QA verification and answer correction, annotators label the candidate viewpoints associated with each question. 
The goal of this step is to identify which viewpoints allow the question to be answered and to remove physically invalid viewpoints.

\paragraph{Viewpoint Annotation Setup.}
For each question, annotators are provided with a set of candidate viewpoint images captured around the target object. 
Using the corrected question and answer as reference, annotators inspect each viewpoint to determine whether the question can be answered from that image.

\paragraph{Answerable Viewpoint Identification.}
Annotators identify the subset of viewpoints from which the question can be correctly answered.
A viewpoint is considered \textit{answerable} if the image provides sufficient visual evidence to determine the answer without relying on additional viewpoints.
If multiple viewpoints satisfy this condition, all of them are marked as answerable.

\paragraph{Invalid Viewpoint Identification.}
Annotators also identify viewpoints that are physically implausible or visually corrupted.
Such viewpoints typically occur when the camera position intersects with scene geometry (e.g., objects, walls, or floors), producing severe rendering artifacts such as fog-like occlusions, or when the viewpoint lies outside the valid indoor space.
These viewpoints are marked as invalid and excluded from consideration.



\subsection{Penetrated Initial Viewpoint Filtering}

Following Step 5 (answerability filtering), we conduct an additional round of human annotation to remove remaining invalid viewpoints. 
Although earlier stages include VLM-based filtering and human annotation, we observe that some invalid viewpoints still remain in the dataset. 

\begin{figure}[h]
\centering
\includegraphics[width=0.9\linewidth]{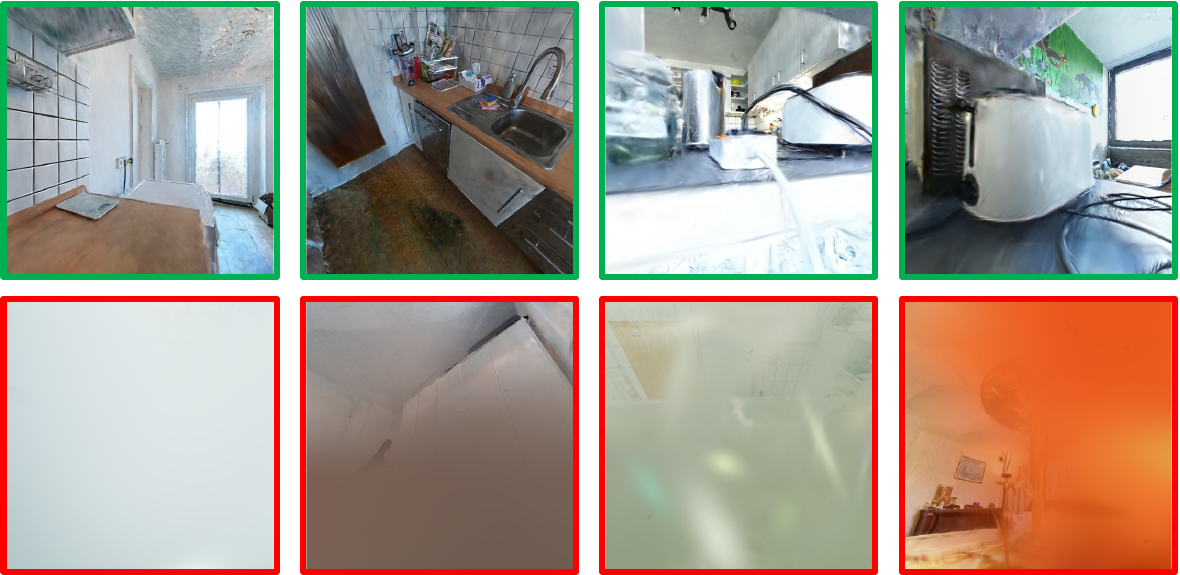}
\caption{
Qualitative results of the penetrated viewpoint filtering.
Viewpoints are marked red if they are filtered out due to camera penetration through scene geometry (e.g., walls or objects). Otherwise, they are marked green to indicate valid initial viewpoints for E3VS.
}
\label{fig:penetrated_viewpoint_filtering}
\end{figure}

Annotators inspect the image corresponding to the agent’s initial position of each episode and remove episodes where the viewpoint is invalid (e.g., penetrated viewpoints). 
Examples of the filtering results are shown in Fig.~\ref{fig:penetrated_viewpoint_filtering}.

\subsection{Counting Modification and Filtering}

Finally, we refine counting questions and remove invalid QA pairs. 
We observe that many questions in the \textit{Counting} category contain viewpoint-dependent expressions that do not reflect the actual object structure in the environment. 
To address this issue, we remove viewpoint-specific phrases such as ``visible'', ``clearly visible'', ``in this view'', and ``in the provided images''. For example:

\begin{itemize}
\item Before: \textit{How many XXX are clearly visible on YYYY?}
\item After: \textit{How many XXX are on YYYY?}
\end{itemize}

This modification reformulates the questions to reflect the structural properties of the target objects rather than a specific viewpoint (Fig.~\ref{fig:prompt_modify_counting}).
Subsequently, we verify the associated answers automatically. If the answers remain viewpoint-dependent or inconsistent with the object structure, the corresponding QA pairs are filtered out. The prompt used for this filtering process is shown in Fig.~\ref{fig:prompt_filter_counting}.

\begin{figure*}[t]
\centering
\begin{minipage}{0.97\linewidth}

\begin{mdframed}[
  backgroundcolor=blue!10,
  linecolor=red!10,
  linewidth=0pt,
  innerleftmargin=10pt,
  innerrightmargin=10pt,
  innertopmargin=6pt,
  innerbottommargin=6pt,
  skipabove=0pt,
  skipbelow=0pt
]
\centering
\bfseries
Prompt for Counting Question Modification
\end{mdframed}

\begin{mdframed}[
  backgroundcolor=blue!2,
  linecolor=white,
  linewidth=0pt,
  innerleftmargin=14pt,
  innerrightmargin=14pt,
  innertopmargin=10pt,
  innerbottommargin=10pt,
  skipabove=0pt,
  skipbelow=0pt
]
\footnotesize

You are a dataset construction assistant for a Navigation + VQA benchmark.

Your task is to minimally edit counting questions so that they require
STRUCTURAL understanding of objects, not surface-level visual counting.

IMPORTANT:
This is a CONSTRAINED, MINIMAL rewrite task.
You must preserve the original question as much as possible.

The rewritten question MUST:
\begin{itemize}
    \item Remove or replace surface-level, view-dependent words (e.g., “visible”, “clearly visible”, “in this view”)
    \item Refer to the total number implied by the object itself
    \item Preserve the original sentence structure and wording as much as possible
    \item Preserve the original counting target exactly
\end{itemize}

You MAY:
\begin{itemize}
    \item Delete unnecessary words or short phrases
    \item Replace words with minimal equivalents (e.g., remove “visible”)
    \item Add very short phrases like “in total” ONLY if strictly necessary
\end{itemize}

You MUST NOT:
\begin{itemize}
    \item Add explanatory phrases such as “based on its structure”
    \item Rephrase the entire sentence
    \item Introduce new objects or attributes
    \item Change the question style or intent
    \item Refer to images, viewpoints, or visibility
\end{itemize}

Prefer deletion over addition.
If a sentence works after removing view-dependent words, do not add anything.

Examples:

Original:
how many visible hinges are there on the white door?
Rewritten:
how many hinges are there on the white door?

Original:
how many distinct horizontal surfaces make up the windowsill in this view?
Rewritten:
how many distinct horizontal surfaces make up the windowsill?

Original:
how many words are clearly visible on the spine of the red book?
Rewritten:
how many words are on the spine of the red book?

Rewrite the following counting question to be structure-based.

Original question: [QUESTION]

Return only the rewritten question.

\end{mdframed}
\end{minipage}
\vspace{-1.0em}
\caption{Prompt for Counting Question Modification}
\vspace{-2.0em}
\label{fig:prompt_modify_counting}
\end{figure*}
\begin{figure*}[t]
\centering
\begin{minipage}{0.97\linewidth}

\begin{mdframed}[
  backgroundcolor=blue!10,
  linecolor=red!10,
  linewidth=0pt,
  innerleftmargin=10pt,
  innerrightmargin=10pt,
  innertopmargin=6pt,
  innerbottommargin=6pt,
  skipabove=0pt,
  skipbelow=0pt
]
\centering
\bfseries
Prompt for Filtering Invalid Counting QAs
\end{mdframed}

\begin{mdframed}[
  backgroundcolor=blue!2,
  linecolor=white,
  linewidth=0pt,
  innerleftmargin=14pt,
  innerrightmargin=14pt,
  innertopmargin=10pt,
  innerbottommargin=10pt,
  skipabove=0pt,
  skipbelow=0pt
]
\footnotesize

You are an expert data curator for a Robot Navigation and Visual Question Answering (VQA) benchmark. 
Your task is to determine if a counting QA pair is 'valid' or 'invalid' based on the relationship between the Image, Question, and Ground Truth (GT) answer.

Core Filtering Logic:
\begin{itemize}
    \item REJECT: The GT count results from partial visibility, occlusion, or camera clipping of a standard object.
  Example: A sofa clearly has 4 legs, but the camera only captures 1, and GT says "1" → INVALID
    \item KEEP: The GT count represents the actual physical structure of the object shown, even if unconventional.
  Example: A tripod or Y-shaped trash bin physically has 3 legs, all visible/accounted for, and GT says "3" → VALID
\end{itemize}

Detailed Criteria:
\begin{enumerate}
    \item Occlusion \& Clipping Check:
    \begin{itemize}
        \item Is the object partially hidden by other furniture or clutter?
        \item If YES and the GT count is very low (1 or 2) compared to the actual number of common parts, it is likely INVALID
    \end{itemize}
    \item Structural Plausibility:
    \begin{itemize}
        \item For the specific object in the image, is the GT count physically plausible for a complete product?
        \item Trash bins, stools, or side tables often have 3 legs → This is VALID
        \item Large sofas, heavy cabinets, or standard dining chairs with only 1-2 legs → INVALID (unless wall-mounted, which is rare)
    \end{itemize}
    \item Ambiguity Check:
    \begin{itemize}
        \item If a human looking at the image would say "There are probably more parts hidden," but GT gives a low count, the task is poor quality → INVALID
    \end{itemize}
\end{enumerate}

Please respond with only 'valid' or 'invalid'.

Question: [QUESTION]
GT Answer: [ANSWER]

\end{mdframed}
\end{minipage}
\vspace{-1.0em}
\caption{Prompt for Filtering Invalid Counting QAs.}
\vspace{-2.0em}
\label{fig:prompt_filter_counting}
\end{figure*}
\section{Implementation Details}
\label{sec:suppl_implementation_details}
We report the inference configurations used to evaluate the VLMs in our benchmark, including decoding settings, reasoning configurations, and output length limits.

\paragraph{Decoding.}
All models are evaluated using deterministic decoding with temperature $=0$, i.e., without sampling. 
However, some models (e.g., \textit{GPT 5.1}) do not allow the temperature to be set to exactly zero via the API. 
For such models, we use the default decoding configuration provided by the API.

\paragraph{Reasoning configuration.}
For \textit{Gemini} models, we control internal reasoning using the parameter thinking\_level. 
Unless otherwise specified, we use thinking\_level=minimal for \textit{Gemini 3.0}, and thinking\_level=medium in ablation experiments requiring explicit reasoning.
For \textit{GPT 5.1}, reasoning is controlled via reasoning\_effort. 
By default, we set reasoning\_effort=none, while the reasoning ablation setting and the VQA module use reasoning\_effort=medium.

\paragraph{Maximum output tokens.}
The maximum number of output tokens is configured depending on the model:

\begin{table}[H]
    \centering
    \renewcommand{\arraystretch}{1.2} 
    \setlength{\tabcolsep}{10pt}      
    \begin{tabular}{lr}
        \toprule
        Model & Max output tokens \\
        \midrule
        \textit{Gemini 2.5 Pro} & 128 \\
        \textit{Gemini 3.0 Flash} & 128 \\
        \textit{Gemini 3.0 Pro} & 256 \\
        \textit{GPT 5.1} & 128 \\
        \textit{Qwen3-VL-8B} & 128 \\
        \textit{Qwen3-VL-30B} & 128 \\
        \textit{InternVL3.5-8B} & 128 \\
        \textit{Step3-VL-10B} & 2,048 \\
        \bottomrule
    \end{tabular}
\end{table}
We use a larger limit for \textit{Step3-VL-10B} because the model explicitly outputs reasoning traces, which substantially increases the token length. 
Using a smaller limit may truncate the output before the final action token is produced, which results in frequent fallback actions (e.g., \texttt{move\_forward}) triggered by our exception handling.
For \textit{Gemini 3.0 Pro}, we set the limit to 256 tokens instead of 128. 
Although the model is instructed to output only discrete actions, it occasionally generates intermediate reasoning text, which may exceed the smaller token limit.

\section{E3VS Framework Details}
\label{sec:suppl_inference_framework_details}


We report the system and user prompts used in the E3VS framework for
VLM-based agents in Fig.~\ref{fig:3dvs_system_prompt} and Fig.~\ref{fig:3dvs_user_prompt}. The system prompt defines the
task setting and the expected response format, while the user prompt
provides the current observation and the question to be answered.
The model is required to produce answers grounded in the visual
evidence of the observation image.

\begin{figure*}[h]
\centering
\begin{minipage}{0.97\linewidth}

\begin{mdframed}[
  backgroundcolor=blue!10,
  linecolor=red!10,
  linewidth=0pt,
  innerleftmargin=10pt,
  innerrightmargin=10pt,
  innertopmargin=6pt,
  innerbottommargin=6pt,
  skipabove=0pt,
  skipbelow=0pt
]
\centering
\bfseries
System Prompt for Embodied 3D Visual Search
\end{mdframed}

\begin{mdframed}[
  backgroundcolor=blue!2,
  linecolor=white,
  linewidth=0pt,
  innerleftmargin=14pt,
  innerrightmargin=14pt,
  innertopmargin=10pt,
  innerbottommargin=10pt,
  skipabove=0pt,
  skipbelow=0pt
]
\footnotesize
 You are an agent controlling a camera in a 3D environment.
 
Your goal is to navigate to the best viewpoint to answer the user's question about the scene.

The world coordinate system has Z as the up-axis.

You will be given your current position, up to 1 recent observation frames (including the current frame). 
The information is shown in chronological order. 
Note: At the initial viewpoint (step 0), the target object is centered in the view. 
However, it may be partially or fully occluded by other objects.

\textbf{Movement Actions:}
Based on the user's question, your state, and the observation(s), choose the next action to reach the best viewpoint.
AVAILABLE ACTIONS: \ttfamily{move\_forward}, \ttfamily{move\_backward}, \ttfamily{move\_left}, \ttfamily{move\_right}, \ttfamily{move\_up}, \ttfamily{move\_down}, \ttfamily{turn\_left}, \ttfamily{turn\_right}, \ttfamily{look\_up}, \ttfamily{look\_down}, \ttfamily{stop}
\begin{itemize}
    \setlength{\itemsep}{0pt}
    \item Each 'move' action translates the camera by 0.25 meters
    \item Each 'turn' or 'look' action rotates it by 30 degrees
    \item Once at an optimal viewpoint, use the 'stop' action
    \item \textbf{Forward/Backward Movement (\ttfamily{move\_forward}, \ttfamily{move\_backward}):} These actions move you along the camera's view direction (camera coordinate axis), which means you will move up or down if the camera is tilted. 
\end{itemize}

\textbf{OUTPUT FORMAT:}
Output exactly one action in this format:
action: \ttfamily{move\_forward}

Example:
action: \ttfamily{move\_forward}

\end{mdframed}
\end{minipage}
\vspace{-1.0em}
\caption{System prompt for Embodied 3D Visual Search.}
\vspace{-2.0em}
\label{fig:3dvs_system_prompt}
\end{figure*}
\begin{figure*}[h]
\centering
\begin{minipage}{0.97\linewidth}

\begin{mdframed}[
  backgroundcolor=blue!10,
  linecolor=red!10,
  linewidth=0pt,
  innerleftmargin=10pt,
  innerrightmargin=10pt,
  innertopmargin=6pt,
  innerbottommargin=6pt,
  skipabove=0pt,
  skipbelow=0pt
]
\centering
\bfseries
User Prompt for Embodied 3D Visual Search
\end{mdframed}

\begin{mdframed}[
  backgroundcolor=blue!2,
  linecolor=white,
  linewidth=0pt,
  innerleftmargin=14pt,
  innerrightmargin=14pt,
  innertopmargin=10pt,
  innerbottommargin=10pt,
  skipabove=0pt,
  skipbelow=0pt
]
\footnotesize

 Question: 'how many legs support the table?'
Current step: 9
Your current state:
\begin{itemize}
    \item Position (X, Y, Z): (1.00, 2.00, 3.00)
    \item Last action: \ttfamily{move\_forward}
    \item Last action result: Collision occurred. You remained at the same position.
\end{itemize}

\end{mdframed}
\end{minipage}
\caption{User prompt for Embodied 3D Visual Search}
\label{fig:3dvs_user_prompt}
\end{figure*}

\section{VLM-as-a-judge Details}
\label{sec:suppl_judge_details}

Since answers in E3VS-Bench are open-vocabulary, exact string matching is inadequate for evaluation. We therefore adopt a VLM-as-a-judge framework.

The judge model receives the question $q$, predicted answer $a_{\text{pred}}$, ground-truth answer $a_{\text{gt}}$, the \textit{goal image} $O_{\text{goal}}$, and the agent’s \textit{end image} $O_{\text{end}}$. The end image corresponds to the final observation used by the agent to generate its answer, enabling the judge to verify whether the prediction is supported by the available visual evidence.

The goal image is additionally provided to account for valid but unannotated answers.
In particular, spatial descriptions often admit multiple correct expressions (e.g., ``next to the chair'' or ``next to the table''). 
In some cases, an instance different from the pre-defined target instance may still serve as a valid target object if it leads to the correct answer. 
To handle such cases, we additionally use the end image for verification.
By referencing the goal image, the judge can resolve such ambiguities and treat semantically correct alternatives as valid.

The judge outputs a score $\sigma \in \{1,2,3,4,5\}$, where 1 indicates an incorrect answer and 5 indicates a correct answer. The full prompt used for the judge model is shown in Fig.~\ref{fig:judge_system_prompt}.

Compared to OpenEQA, where ground-truth answers typically consist of short phrases or single words, the answers in E3VS-Bench are often longer and contain more descriptive expressions.
As a result, semantically correct predictions may exhibit greater lexical variation with respect to the reference answers.
This characteristic makes exact or near-exact matching more difficult and can naturally reduce the correlation between automated judging and human evaluation.
In addition, our evaluation protocol requires the judge to verify whether the predicted answer is supported by the available visual evidence in the end image, which further increases the difficulty of automated assessment.
Despite these challenges, we observe a moderate correlation (Spearman's $\rho = 0.54$), indicating that the VLM-based judge still provides a reasonable proxy for human assessment.

We additionally evaluated a stronger judge model, \textit{GPT 5.5}, and
observed an improved correlation of $\rho = 0.615$.
This result suggests that the reliability of automatic evaluation can
further benefit from advances in multimodal foundation models.




\FloatBarrier 
\begin{tcolorbox}[
    breakable,       
    enhanced,
    title=System Prompt for VLM-as-a-Judge,
    fonttitle=\bfseries\centering,
    colback=blue!2,           
    colbacktitle=blue!10,     
    coltitle=black,           
    colframe=red!10,          
    sharp corners,
    boxrule=0.5pt,
    left=10pt, right=10pt, top=10pt, bottom=10pt,
    before skip=10pt, after skip=10pt
]
\scriptsize

You are a discerning and fair judge evaluating a Vision-Language Model's performance in a Navigation + VQA task.
Your goal is to determine if the Model's RESPONSE provides a visually grounded and correct answer to the QUESTION.

Input Data:
\begin{enumerate}
    \setlength{\itemsep}{0pt}
    \item END\_IMAGE: The actual view seen by the model at the end of navigation.
    \item GOAL\_IMAGE: The image used to derive the ground-truth answer.
    \item QUESTION: The query asked.
    \item RESPONSE: The model's answer.
    \item REFERENCE INFO: One example of a correct answer.
\end{enumerate}

\nointerlineskip \vspace{.5em}
\hrule
\vspace{.5em}

\noindent{\normalsize  Core Principles}
\begin{itemize}
    \setlength{\itemsep}{0pt}  
    \item Visual Grounding is Paramount: The answer MUST be derived from the END\_IMAGE.
    \item No Blind Guessing: Even if the RESPONSE matches the REFERENCE INFO, it is incorrect if the object is NOT visible in the END\_IMAGE.
    \item Object Existence: An object is considered to exist only if it is clearly visible in the END\_IMAGE. Do not accept assumptions based on room layout.
    \item Reference Semantics: The REFERENCE INFO and GOAL\_IMAGE define the type of object expected. The RESPONSE must match this semantic identity (e.g., if Reference is "Chair", "Table" is wrong, even if a table is visible).
\end{itemize}

\nointerlineskip \vspace{.5em}
\hrule
\vspace{.5em}

\noindent{\normalsize  Evaluation Logic}

Step 1: False Negative Check (Strict)
\begin{itemize}
    \setlength{\itemsep}{0pt}  
    \item If the RESPONSE denies the possibility of answering (e.g., "not possible", "cannot find", "nowhere", "I don't see"), while the REFERENCE INFO indicates a valid answer: → Score 1
\end{itemize}

Step 2: Spatial Dimensionality Check (Strict for "Where" questions)
\begin{itemize}
    \setlength{\itemsep}{0pt}  
    \item If the QUESTION asks for a location:
    \begin{itemize}
        \item Does the RESPONSE describe the position using 2D image coordinates (e.g., "right side of the image", "bottom corner")?
        \item Instead of 3D spatial relationships (e.g., "on the table", "next to the chair")?
    \end{itemize}
    \item If yes (uses 2D only): → Score 1
\end{itemize}

Step 3: Object Existence \& Identity Validation (Critical)

\begin{itemize}
    \setlength{\itemsep}{0pt}  
    \item Identify all objects referenced in the RESPONSE.
    \item For each referenced object:
    \begin{enumerate}
        \item Visibility: Is the object clearly and unambiguously visible in the END\_IMAGE? (Reject hallucinations or objects inferred from dark/blurry areas).
        \item Identity: Does the visual appearance in END\_IMAGE match the object name in the RESPONSE?
        \begin{itemize}
            \item Example Failure: Calling a "Mop" a "Toilet Brush".
            \item Example Failure: Calling a "Table" a "Chair".
            \item Use GOAL\_IMAGE to resolve ambiguity about what the target object looks like.
        \end{itemize}
    \end{enumerate}
    \item If any referenced object is invisible or misidentified (wrong category): → Score 1
\end{itemize}


Step 4: Visual \& Spatial Consistency Check

\begin{itemize}
    \setlength{\itemsep}{0pt}  
    \item Verify the spatial relationship (e.g., "on", "under", "next to").
    \item Strict Preposition Check:
    \begin{itemize}
        \item "Next to" is incorrect if the object is "On" a surface.
        \item "Under" is incorrect if the object is "Next to" or "In front of".
    \end{itemize}
    \item If the spatial description implies a relationship not supported by the END\_IMAGE pixels: → Score 1
\end{itemize}


Step 5: Final Verdict \& Reference Comparison

Case A: RESPONSE matches REFERENCE INFO (Word-for-word or exact meaning)
\begin{itemize}
    \setlength{\itemsep}{0pt}  
    \item CRITICAL VERIFICATION: Do NOT automatically award 5.
    \item Check theEND\_IMAGE: Is the object/location described actually visible in the current view?
    \begin{itemize}
        \setlength{\itemsep}{0pt}  
        \item If the object is occluded, out of frame, too dark, or invisible in the END\_IMAGE (implying the model guessed correctly by chance or dataset bias): → Score 1 (Blind Guessing)
        \item If the REFERENCE INFO itself seems wrong based on the END\_IMAGE (annotation error) and the model followed it blindly against visual evidence: → Score 1
        \item If the object is visible and the answer is visually correct: → Score 5
    \end{itemize}
\end{itemize}

Case B: RESPONSE does NOT match REFERENCE INFO
\begin{itemize}
    \setlength{\itemsep}{0pt}  
    \item The RESPONSE is a Valid Alternative (Score 5) ONLY if:
    \begin{enumerate}
        \setlength{\itemsep}{0pt}
        \item It refers to objects validated in Step 3 (Clearly visible).
        \item It satisfies the same semantic intent as the REFERENCE/GOAL.
        \begin{itemize}
            \item Acceptable: Synonyms ("Trash can" vs "Bin"), Shape approximations ("Circular" vs "Oval"), Hypernyms ("Furniture" vs "Chair" - if accurate).
            \item Unacceptable: Different object categories ("Table" vs "Chair"), contradicting attributes ("Red" vs "Blue").
        \end{itemize}
        \item It is visually consistent with the END\_IMAGE.
    \end{enumerate}
    \item If it fails any of these: → Score 1
    \item If it is a valid alternative: → Score 5
\end{itemize}

\nointerlineskip \vspace{.5em}
\hrule
\vspace{.5em}

\noindent{\normalsize Output Format}

Reasoning: [Explain the score based on the steps above. Explicitly mention if the object was visible in END\_IMAGE and if the Identity matched.]
Your mark: [Integer from 1 to 5]

\end{tcolorbox}
\nopagebreak
\captionof{figure}{System Prompt for VLM-as-a-Judge}
\label{fig:judge_system_prompt}

\FloatBarrier 

\end{document}